\documentclass{article}

\usepackage[margin=1.0in]{geometry}
\usepackage[utf8]{inputenc} 
\usepackage[T1]{fontenc}    
\usepackage{url}            
\usepackage{booktabs}       
\usepackage{amsfonts}       
\usepackage{nicefrac}       
\usepackage{microtype}      
\usepackage{xcolor}
\usepackage{lineno}
\usepackage{caption}
\usepackage{multirow}
\usepackage{setspace}
\usepackage{subcaption}
\usepackage{xr-hyper}
\makeatletter
\newcommand*{\addFileDependency}[1]{
  \typeout{(#1)}
  \@addtofilelist{#1}
  \IfFileExists{#1}{}{\typeout{No file #1.}}
}
\makeatother

\usepackage{graphicx}
\usepackage[style=nature]{biblatex}
\usepackage{algorithm}
\usepackage[noend]{algpseudocode}
\usepackage{amsmath}
\usepackage{amsthm}
\usepackage{enumitem}
\usepackage{authblk}
\algnewcommand{\IIf}[1]{\State\algorithmicif\ #1\ \algorithmicthen}
\algnewcommand{\EndIIf}{\unskip\ \algorithmicend\ \algorithmicif}

\usepackage{todonotes}
\usepackage{datenumber}
\usepackage{eqname}
\usepackage{float}
\usepackage{setspace}

\title{Forecasting adverse surgical events using self-supervised transfer learning for physiological signals}

\author[1]{Hugh Chen}
\author[2]{Scott M. Lundberg}
\author[1,3]{Gabriel Erion}
\author[4]{Jerry H. Kim}
\author[1,*]{Su-In Lee}

\affil[1]{{\small Paul G. Allen School of Computer Science and Engineering, University of Washington}}
\affil[2]{{\small Microsoft Research}}
\affil[3]{{\small Medical Scientist Training Program, University of Washington}}
\affil[4]{{\small Global Innovation Exchange, University of Washington}}
\affil[*]{{\small Corresponding: suinlee@cs.washington.edu}}

\begin{document}

\setcounter{page}{1}

\date{}

{\setstretch{1}
\maketitle
}

\begin{abstract}
Hundreds of millions of surgical procedures take place annually across the world,  
which generate a prevalent type of electronic health record (EHR) data comprising time series physiological signals. Here, we present a transferable embedding method (i.e., a method to transform time series signals into input features for predictive machine learning models) named PHASE (PHysiologicAl Signal Embeddings) that enables us to more accurately forecast adverse surgical outcomes based on physiological signals 
We evaluate PHASE on minute-by-minute EHR data of more than 50,000 surgeries from two operating room (OR) datasets and patient stays in an intensive care unit (ICU) dataset.  PHASE outperforms other state-of-the-art approaches, such as long-short term memory networks trained on raw data and gradient boosted trees trained on handcrafted features, in predicting five distinct outcomes: hypoxemia, hypocapnia, hypotension, hypertension, and phenylephrine administration.  In a transfer learning setting where we train embedding models in one dataset then embed signals and predict adverse events in unseen data, PHASE achieves significantly higher prediction accuracy at lower computational cost compared to conventional approaches.  Finally, given the importance of understanding models in clinical applications we demonstrate that PHASE is explainable and validate our predictive models using local feature attribution methods.
\end{abstract}

\section{Introduction}


Globally, the number of surgical operations performed each year exceeds 300 million \cite{weiser2016size}.  Although surgeries are crucial components of medical care, they have a higher prevalence of adverse events (i.e., patients harmed as a result of their medical treatment) relative to other medical specialties.  In fact, several international studies have shown rates of adverse events ranging from 3\% to 22\% in surgical patients \cite{nilsson2016preventable,zegers2011incidence,kable2002adverse}.  Fortunately, these studies also conclude that the majority of adverse events are preventable, indicating a tremendous opportunity for improvement by predictive models.



The accuracy of such models is largely dependent on the availability of training data.  As of 2014, a large portion ($>40\%$) of invasive, therapeutic surgeries take place in hospitals with either medium or small numbers of beds \cite{steiner2017,wen2016}.  These smaller institutions may lack either sufficient data or computational resources to train accurate models.  Furthermore, patient privacy considerations mean that large public EHR datasets are unlikely, leaving many institutions with insufficient resources to train performant models on their own.  In the face of this insufficiency, one natural way to make accurate predictions is \emph{transfer learning}, which has already shown success in medical images as well as clinical text \cite{7426826,ravishankar2016understanding,LV201455}.  Particularly with the popularization of wearable sensors for health monitoring \cite{majumder2017wearable}, transfer learning techniques that train models in one dataset and use them in another are arguably underexplored for physiological signals, which account for a significant portion of the hundreds of petabytes of currently available worldwide health data \cite{roski2014creating,orphanidou2019review}.  One promising avenue of transfer learning research is \emph{deep embedding models} which learn to extract generalizable features from images or time series data \cite{chen2016deep,malhotra2017timenet} which improve over traditional domain-specific hand engineered features.


Our approach, PHASE (PHysiologicAl Signal Embeddings), trains deep embedding models on physiological signals to better forecast and facilitate prevention of potentially millions of adverse surgical outcomes.  Furthermore, these models not only improve predictive accuracy but can be transferred from an institution with plentiful computational resources to institutions with less.  PHASE improves over previous approaches in two important ways: 

\begin{itemize}
    \item PHASE \textit{improves predictive accuracy} by leveraging deep learning to embed physiological signals.   Using long-short term memory networks (LSTMs), PHASE embeds physiological signals prior to forecasting adverse events with a downstream model.  We investigate a number of self-supervised approaches (training with inputs and outputs derived from the signal data itself) \cite{kolesnikov2019revisiting} to effectively train embedding models.  Our results show that gradient boosted tree (GBT) models trained with features extracted by self-supervised LSTMs improves accuracy over conventional approaches for forecasting surgical outcomes that rely on a single model (i.e., predicting adverse outcomes with an LSTM with raw features or a GBT with raw or hand engineered features).
    \item PHASE \textit{shares models rather than data} to address data insufficiency and improves over alternative methods including GBTs trained with raw features, hand engineered features, and embeddings jointly learned by a single LSTM.  Data insufficiency is especially important for surgical data because protecting patient privacy makes it difficult to share large amounts of medical data which exacerbates the lack of publicly available data \cite{kohli2017medical}.  By transferring performant models as has been done in medical images and clinical text \cite{7426826,ravishankar2016understanding,LV201455}, scientists can collaborate to improve accuracy of predictive models without exposing patient data.
\end{itemize}



\begin{figure}
    \centering
    \includegraphics[width=.8\textwidth]{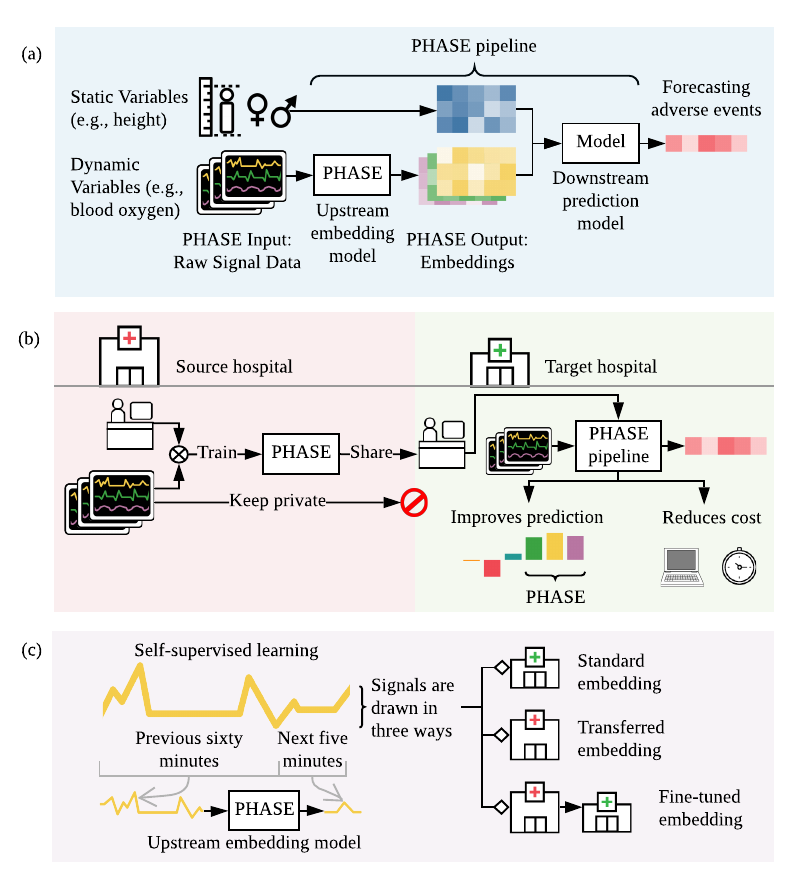}
    \caption{
    (a) PHASE learns models that embed (i.e., extract features from) physiological signals. We concatenate these embeddings with static data to predict adverse events.  We describe the model extracting features as an \textit{upstream embedding model} and the model making the final prediction as the \textit{downstream prediction model}.  (b) PHASE enables researchers at different hospitals to work together without sharing data.  Researchers can perform transfer learning where upstream embedding models are trained on data drawn from a \textit{source hospital} and used to embed signals and make a downstream prediction in data drawn from a \textit{target hospital}. We show that this approach outperforms conventional deep learning and tree models trained with raw or hand engineered features.  In addition, this approach reduces computational cost for users in target hospitals. (c) PHASE comprises LSTM embedding models trained per physiological signal that predict the future of the signal based on the past (self-supervised learning).  We train self-supervised embedding models using data drawn in three distinct ways: (1) from the target hospital (standard embedding), (2) from a distinct source hospital (transferred embedding), and (3) from a distinct source hospital and then the target hospital (fine-tuned embedding) (More details in Section \ref{sec:overview}).}
    \label{fig:concept}
\end{figure}


In contrast to prior research on transfer learning for physiological signals that focus on a single medical center's electroencephalograms (EEGs) \cite{fahimi2018inter} or intensive care unit (ICU) stays \cite{gupta2019transfer}, we evaluate transfer learning across three distinct medical center data sets (two from operating rooms and one from an ICU).  Furthermore, we focus on evaluating self-supervised approaches (Figure \ref{fig:concept}) to train embedding models that we validate with feature attributions. To achieve this, we use data collected by the Anesthesia Information Management System (AIMS) from two medical centers as well as the Medical Information Mart for Intensive Care (MIMIC-III) dataset \cite{johnson2016}.   We utilize fifteen physiological signal variables and six static variable inputs (variables listed in Section \ref{sec:outcomes}) to forecast five possible outcomes: hypoxemia, hypocapnia, hypotension, hypertension, and phenylephrine administration.
We show in a standard embedding setting, PHASE outperforms a number of conventional approaches across five outcomes of interest: hypoxemia, hypocapnia, hypotension, hypertension, and phenylephrine administration.  Our results suggest that if the previous state of the art machine learning model (a gradient boosted tree model using hand engineered features \cite{lundberg2018explainable}) captured 15\% of hypoxemic events, PHASE captures approximately 19\% of hypoxemic events based on a fixed recall.   In our dataset we observe approximately 2.3 hypoxemic events per surgery, in the US alone our method could forecast roughly 1 million hypoxemic events that the previous state of the art model fails to capture (given that there are an annual 10 million surgeries in the US).

Furthermore, we show that PHASE improves performance in a transferred embedding setting where LSTM embedding models are trained in one dataset and used to extract features in a completely unseen dataset.  Building upon this finding, we show that fine-tuning the LSTMs on unseen data leads to faster convergence and improved predictive performance compared to randomly initialized models across all outcomes.  Finally, we validate our models by identifying important variables using state of the art local feature attribution methods \cite{lundberg2020local}.  We interpret our models to validate that the models uncover statistical patterns that agree with prior literature and demonstrate that models trained using PHASE are explainable.  Importantly, explainability ensures that models are fair, trustworthy, and valuable to scientific understanding \cite{doshi2017towards}.  PHASE takes a step in the direction of allowing scientists to collaborate on EHR data which is typically accessible by only a single group (data silos \cite{rai2016risk}) by investigating approaches to train embedding models that generalize to unseen data.

\section{Results}

\subsection{Five perioperative outcomes from three hospital datasets}
\label{sec:outcomes}

We are interested in forecasting five outcomes associated with surgical morbidity.  The first is hypoxemia (i.e., low blood oxygen level), an important cause of anesthesia-related morbidity, resulting in
harmful effects on nearly every end organ 
\cite{korner1959circulatory,ehrenfeld2010incidence}.  The next three outcomes are hypocapnia (i.e., low blood carbon dioxide), hypotension (i.e., low blood pressure), and hypertension (high blood pressure).  Negative physiological effects associated with hypocapnia include reduced cerebral blood flow and reduced cardiac output \cite{pollard1977some}.  Prolonged episodes of perioperative hypotension are associated with end-organ ischemia as well as assorted other adverse postoperative complications \cite{lienhart2006survey,chang2000adverse,jeremitsky2003harbingers}.  
In addition, perioperative hypertension has been tied to increased risk of postoperative intracranial hemorrhage in craniotomies \cite{basali2000relation} and end organ dysfunction \cite{varon2008perioperative}.  Finally, the last outcome of interest is the administration of phenylephrine, a medication frequently used to treat hypotension during anesthesia administration \cite{kee2004prophylactic}.  
Predicting phenylephrine use lets us further evaluate PHASE because it represents a clinical decision rather than an aspect of patient physiology as in the prior four outcomes.



\begin{table}[!ht]
\centering
\begin{tabular}{l|l|ccc}
& \textbf{Dataset}                         & OR$_0$ & OR$_1$ & ICU$_\text{M}$  \\
\hline
\hline
& Department & OR & OR & ICU \\ 
& Number of procedures/stays & 29,035 & 28,136 & 1,669 \\ \hline
Static & Gender (\% female) & 57\% & 38\% & 44\% \\ 
variables & Age (yr) Mean & 51.859 & 48.701 & 63.956 \\
& Age (yr) Std. & 16.748 & 18.419 & 17.708 \\
& Weight (lb) Mean & 185.273 & 181.608 & 176.662 \\
& Weight (lb) Std. & 54.042 & 54.194 & 55.448 \\
& Height (in) Mean & 66.913 & 67.502 & 66.967 \\
& Height (in) Std. & 8.268 & 8.607 & 6.181 \\
& ASA Code Emergency & 7.65\% & 15.31\% & - \\ \hline
Adverse & Hypoxemia Base Rate & 1.09\% & 2.19\% & 3.93\%\\
outcomes & Hypocapnia Base Rate & 9.76\% & 8.06\% & -\\
& Hypotension Base Rate & 7.44\% & 3.53\% & -\\
& Hypertension Base Rate & 1.70\% & 1.66\% & -\\
& Phenylephrine Base Rate & 7.23\% & 9.15\% & -\\
\end{tabular}
\caption{Training set statistics for different data sources.  Each outcome has a different number of samples due to missing data.}
\label{tab:hospitals}
\end{table}

To evaluate our methodology with these outcomes, we utilize data from three different hospital datasets, summarized in Table \ref{tab:hospitals} (Methods Section \ref{sec:data_cohort} and Supplementary Section \ref{sec:app:data}).  In brief, we consider two operating room datasets from distinct medical centers which we denote as OR$_0$ and OR$_1$.  We also use the publicly available intensive care unit MIMIC-III dataset which we refer to as ICU$_\text{M}$ \cite{johnson2016}.  As inputs, we use fifteen physiological signal variables: {\it SAO2} - Blood oxygen saturation, {\it ETCO2} - End-tidal carbon dioxide, {\it NIBP[S/M/D]} - Non-invasive blood pressure (systolic, mean,  diastolic), {\it FIO2} - Fraction of inspired oxygen, {\it ETSEV/ETSEVO} - End-tidal sevoflurane, {\it ECGRATE} - Heart rate from ECG, {\it PEAK} - Peak ventilator pressure, {\it PEEP} - Positive end-expiratory pressure, {\it PIP} - Peak inspiratory pressure, {\it RESPRATE} - Respiration rate, {\it TEMP1} - Body temperature in addition to six static variables: {\it Height}, {\it Weight}, {\it ASA Code}, {\it ASA Code Emergency}, {\it Gender}, and {\it Age}.  All variables are consistently measured in the operating room datasets, but only {\it SAO2} is consistently measured in the ICU dataset.

Our metric of evaluation is the area under a precision recall curve, otherwise known as average precision (AP), which is more informative than the area under a receiver operating curve (ROC AUC) for binary predictions with low base rates \cite{davis2006relationship}, as in the outcomes we consider.  In particular, we focus on the percent improvement over using the raw, unprocessed physiological signals as an evaluation metric, which is analogous to transfer loss: the difference between the transfer error and the in-domain baseline error \cite{glorot2011domain}.  We additionally report the absolute value of the AP (and ROC AUC for a subset of results) in Supplementary Section \ref{sec:app:absolute_AP}.

\subsection{Overview of the PHASE framework}
\label{sec:overview}
PHASE is an approach to embed physiological signals.  We consider an embedding framework using \emph{upstream embedding models} $U$ that are trained for each physiological signal in a source hospital data set $H_s$.  We evaluate upstream embedding models with a downstream prediction model $D$ whose inputs are the embedded physiological signals concatenated to static variables and outputs are adverse surgical outcomes.  $D$ is trained in a target hospital data set $H_t$.  We evaluate our models in three ways (Figure \ref{fig:concept}c): (1) standard embedding where the source hospital is the same as the target hospital $H_s=H_t$ (Figures \ref{fig:performance}b and \ref{fig:performance}d), (2) transferred embedding where the source hospital is different to the target hospital $H_s\neq H_t$ (Figures \ref{fig:performance}c and \ref{fig:performance}d), and (3) fine-tuned embedding where the upstream embedding model is first trained to convergence in a different source hospital $H_s\neq H_t$ and then used to initialize a model that is trained to convergence in the target hospital $H_s=H_t$ (Figure \ref{fig:ft_performance}).  

The modeling decision of \textit{per-signal} upstream embedding was driven by several advantages: (1) we showed that per-signal embedding models produce embeddings that outperform downstream prediction models trained on the raw signals or hand-engineered signal features (described in Section 2.4) (2) we found that per-signal embedding models worked better than a single embedding model trained on all signals jointly in (Supplementary Section \ref{sec:app:joint}), and (3) we demonstrate that per-signal embedding models work even in a heterogeneous setting where the variables available in the target hospital are different to the variables available in the source hospital (Supplementary Section \ref{sec:app:heterogeneous}).

Here, we briefly describe the embeddings: \textit{raw, ema, rand, auto, next, min}, and \textit{hypo} in \ref{fig:performance}a (more details in Methods Section \ref{sec:meth:features}).  \textit{Raw} and \textit{ema} are not deep learning models.  Instead, \textit{raw} is the raw signal itself and \textit{ema} are exponential moving averages and variance features from \citeauthor{lundberg2018explainable} \cite{lundberg2018explainable}.  The remaining embeddings all use the final hidden layer of LSTMs trained in a source hospital $H_s$ to embed the signals.  The first embedding is \textit{rand}, which uses an untrained LSTM with random weights.  The second is an unsupervised approach called \textit{auto}, which uses an LSTM trained to autoencode the input.  The following two approaches (\textit{next} and \textit{min}) are self-supervised: the LSTM outputs are drawn from the same physiological signal variable as the input, but are taken from different parts of the signal.  \textit{Next} uses LSTMs trained to predict the next five minutes of a particular signal; \textit{min} uses LSTMs trained to predict the minimum of the next five minutes of a particular signal. The final approach, \textit{hypo}, is a traditional supervised approach to transfer learning where the embedding model has the same output as the downstream prediction model (either hypoxemia, hypocapnia, or hypotension).

\subsection{Comparing approaches to embed physiological signals}
\label{sec:results:performance}

\begin{figure}
    \centering
    \includegraphics[width=.9\textwidth]{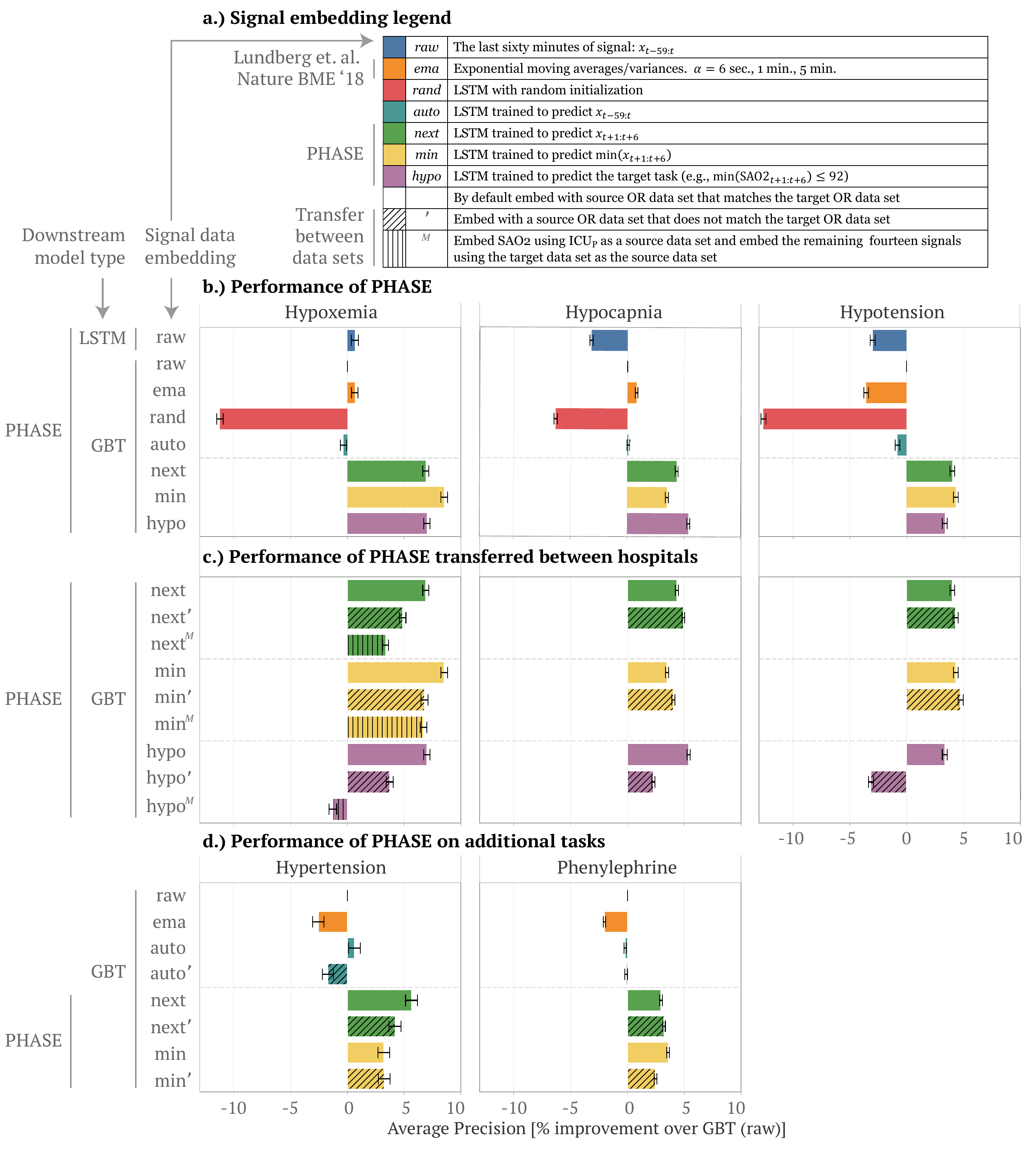}
    \caption{Comparing the performance of downstream models trained with different embeddings of physiological signals concatenated to static features.  We report the average precision (\% improvement over GBT model trained with \textit{raw} signal data, 99\% confidence intervals from bootstrapping the test set).  We use OR$_0$ and OR$_1$ as target datasets and then aggregate across both by averaging the resultant means and standard errors of the \% improvement.  (a) The upstream embedding models we use to extract the physiological signal features where raw is the identify function, ema is an exponential moving average, and the rest are LSTMs trained in specific ways. (b) The performance of downstream prediction models for a variety of standard embedding approaches (when the source hospital is the same as the target hospital).  We compare combinations of downstream models and embeddings for three adverse surgical outcomes (hypoxemia, hypocapnia, and hypotension).  (c) The performance of transferred embedding (\textit{next}’, \textit{next}$^M$, \textit{min}’, \textit{min}$^M$, \textit{hypo}’, and \textit{hypo}$^M$) vs. non-transferred (\textit{next}, \textit{min}, and \textit{hypo}) models for the above three adverse outcomes.  In the transferred approaches the source hospital is different to the target hospital.  (d) Performance of approaches for standard and transferred embedding on additional outcomes: hypertension (high, rather than low, blood pressure) and phenylephrine (doctor action prediction). We do not evaluate \textit{hypo} embeddings in this setting, because the outcomes are not ``hypo'' events.  Model architectures in Supplementary Section \ref{sec:app:arch}.}
    \label{fig:performance}
\end{figure}

As a start, we first compare two popular machine learning models (GBTs and LSTMs) trained on the raw signal data (i.e., without embedding) concatenated to static patient data.  In this section we will refer to results according to (1) the downstream model type and (2) the signal embedding type (for instance, GBT \textit{raw} denotes a gradient boosted tree model trained with the raw minute by minute signal data).  In Figure \ref{fig:performance}b, GBT \textit{raw} performs comparably to LSTM \textit{raw} for hypoxemia and better for hypocapnia and hypotension even though the LSTM should be more suitable to the time series signal data.  Based on prior literature, we hypothesize that the GBT better captures patterns in the static patient data which is tabular \cite{lundberg2020local}, but the LSTMs better capture patterns in the time series data.  In order to leverage the advantages of both model types, we propose PHASE which utilizes LSTMs to embed physiological signals and GBTs to perform the final prediction using the extracted features concatentated to static patient data (Figure \ref{fig:concept}a).  In the following sections we primarily use GBTs as the downstream model and when we refer to our results solely by the signal data embedding they are assumed to use GBTs as the downstream model (for instance, \textit{next} denotes a GBT model trained with \textit{next} embedded data).

We first evaluate the PHASE methods that include two self-supervised embeddings (\textit{next} and \textit{min}) and a supervised embedding (\textit{hypo}) in a standard embedding setting where the source dataset is the same as the target dataset (Figure \ref{fig:performance}b). 
We train GBT downstream models on the physiological signal embeddings concatenated to static patient features to see if the embeddings are more informative than the raw signals.  \textit{Rand} (which serves as a lower bound) transforms physiological signals in an uninformative manner and makes it harder to predict the outcomes of interest in comparison to the raw signals.  Furthermore, \textit{ema} and \textit{auto} fail to consistently improve or impair performance relative to \textit{raw} and thus are not viable features.  In contrast, the PHASE methods (\textit{next}, \textit{min}, and \textit{hypo}) consistently yield models that outperform the alternative approaches across all three outcomes (all p-values $<0.05$). In particular, \textit{ema} is a gradient boosted tree model trained with hand engineered features (exponential moving average) shown to be on par with practicing anesthesiologists at forecasting hypoxemia (Lundberg et. al. Nature BME 2018 \cite{lundberg2018explainable}).  PHASE embeddings further improve over this approach suggesting that PHASE outperforms clinicians for forecasting hypoxemia by approximately 5\% (Figure \ref{fig:performance}b). 

In order to see how the choice of embedding model output affects downstream model performance we can take a closer look at \textit{auto}, \textit{next}, \textit{min}, and \textit{hypo}.  Contrasting PHASE embeddings to \textit{auto} suggests that \textit{incorporating the future in the source task is crucial} (as in \textit{next}, \textit{min}, and \textit{hypo}). However, while taking the minimum (\textit{min}) and thresholding (\textit{hypo}) make the upstream embedding model's outcome more similar to the downstream prediction model's outcome, \textit{min} and \textit{hypo} embeddings do not consistently improve downstream prediction performance compared to \textit{next}.

The previously described results show that PHASE works when forecasting hypoxemia, hypocapnia, and hypotension; however these outcomes are all associated with low signals (hence the ``hypo'' prefix).  In order to validate that PHASE performs well for ``non-hypo'' outcomes as well, we consider two additional outcomes: hypertension (i.e., high blood pressure) and phenylephrine administration (doctor action prediction) (Figure \ref{fig:performance}d).  For hypertension we empirically demonstrate that \textit{next} embeddings are better then \textit{min} embeddings.  This is to be expected because \textit{min} focuses on the minimum of the future signal, whereas hypertension is defined as blood pressure being too high and it therefore addresses the maximum of the future signal.  For phenylephrine, both the \textit{next} and \textit{min} models improve over standard approaches.  One potential reason is that phenylephrine is typically administered in response to low blood pressure and thus \textit{min} models are relevant to phenylephrine administration. 

\subsection{Evaluating upstream embedding models on unseen data}
\label{sec:results:transference}

Previously we focused on a standard embedding setting in a single medical center; in this section, we examine the performance of PHASE when the upstream LSTM embedding models are trained in one dataset but used to embed signals in an unseen dataset (i.e., \emph{transferred} embedding setting).  We analyze two distinct transfer learning settings where the source hospital differs to the target hospital (more details in Methods Section \ref{sec:meth:transferlearning}).  We utilize a superscript notation ($'$ and $^M$) to denote transfer learning.  The apostrophe ($'$) denotes that we trained LSTMs in one operating room dataset and then fixed them to embed signal variables and evaluate performance with a downstream GBT model in the other.  The superscript M ($^M$) denotes that we trained the LSTM for SAO2 in ICU$_\text{M}$ and the other LSTMs in the target dataset\footnote{MIMIC-III (ICU$_\text{m}$) has high rates of missingness for signals except for ECG (which is not directly present in the OR datasets) and SAO2.  This means we were able to train an upstream LSTM only for SAO2 from ICU$_\text{m}$ and we extracted features from the remaining signals using LSTMs trained in the target domain.  This result is still meaningful, because it means we can use upstream embedding models trained in different domains synergistically.}.




Training the LSTM embedding models on a source dataset that differs from the target dataset and using a GBT downstream model ($'$ and $^M$ in Figures \ref{fig:performance}c and \ref{fig:performance}d) generally outperforms conventional approaches: the LSTM trained on raw data and the GBT trained on raw or engineered features (LSTM \textit{raw}, GBT \textit{raw}, and \textit{ema} in Figures \ref{fig:performance}b and \ref{fig:performance}d).  The \textit{next} and \textit{min} embeddings in the transferred embedding settings (\textit{next}$'$, \textit{min}$'$, \textit{next}$^M$, \textit{min}$^M$) outperform the conventional approaches for all possible outcomes (Figure \ref{fig:performance}c) including hypertension and phenylephrine (Figure \ref{fig:performance}d).  However, for \textit{hypo}, the supervised embedding, \textit{hypo}$'$ improves over \textit{raw} embeddings for hypoxemia and hypocapnia, but actually hurts performance for hypotension.  Furthermore the \textit{hypo}$^M$ embedding also hurts performance for hypoxemia relative to using the \textit{raw} embedding.  This suggests that the choice of LSTM embedding model output is important and the supervised learning outcome (\textit{hypo}$'$, \textit{hypo}$^M$) does not generalize to unseen data as well as the self-supervised approaches (\textit{next}$'$, \textit{next}$^M$, \textit{min}$'$, \textit{min}$^M$).


Comparing the transferred embedding models ($'$ and $^M$ in Figure \ref{fig:performance}c and \ref{fig:performance}d) to the standard embedding models (\textit{next}, \textit{min}, \textit{hypo} in Figures \ref{fig:performance}c and \ref{fig:performance}d) we see that the transferred embedding models generally perform comparably to the standard embedding models even though they are evaluated on previously unseen data.  In particular, we see that the \textit{next}$'$, \textit{min}$'$, \textit{next}$^M$, and \textit{min}$^M$ embeddings perform comparably to their standard, non-transferred counterparts (\textit{next} and \textit{min}).  It is worth noting that the transferred embeddings are equally performant for hypocapnia and hypotension; however, slightly reduce downstream performance for hypoxemia and hypertension, which may be due to differences in the hospital data sets (e.g., covariate shift).  As before, we see that the \textit{hypo}$'$ and \textit{hypo}$^M$ embeddings perform substantially worse than their non-transferred counterpart \textit{hypo}.

Although transferred PHASE embeddings perform slightly worse in the hypoxemia and hypertension prediction settings, one important advantage of transferring models is that end users in the target domain can use them at \textit{no additional training cost}.  Training all upstream LSTM embedding models for \textit{next} constituted roughly 66 hours on an NVIDIA GeForce RTX 2080 Ti GPU.   Clinicians who lack either computational resources or deep learning expertise to train their own models from scratch can instead use an off-the-shelf, fixed embedding model.  Given that machine learning is usually not the primary concern of hospital staff, fixed embedding models are a straightforward way to improve the performance of models trained on physiological signal data at minimal cost to the end users.  


There are two additional considerations for transfer learning: (1) In our results, we focus on evaluation using GBT downstream models.  In order to show that the features we extract consistently boost performance and are robust to the choice of the downstream model we replicate our results for a multilayer perceptron (MLP) downstream model in Supplementary Section \ref{sec:app:mlp_performance}.  (2) Per-signal LSTM embedding models outperform a single LSTM embedding model jointly trained with all signals in Supplementary Section \ref{sec:app:joint}.  However, per-signal embedding models have an additional advantage: they work even when the variables available in the target hospital do not exactly match the ones in the source hospital (\textit{feature heterogeneity}).  Per-signal LSTM embedding models work in heterogeneous settings because end users can pick and choose models that correspond to the signals available at their institution.  In comparison, a model trained on all possible variables would be unusable on a new hospital dataset with different variables.  In Supplementary Section \ref{sec:app:heterogeneous}, we show that in heterogeneous settings where the target hospital has fewer features than the source hospital, GBTs trained with PHASE consistently outperform GBTs trained with the raw signals.

\subsection{Fine-tuning upstream embedding models improves performance and reduces computational cost}
\label{sec:results:fine_tune}


In Section \ref{sec:results:transference} we discussed that using PHASE embedding models in the transferred embedding setting are preferable to the standard embedding setting in terms of training cost; however, the standard embedding models still showed slightly better performance for hypoxemia and hypertension.  Alternatively, we propose a fine-tuned embedding approach where we assume an end user in the target hospital has been provided a pre-trained embedding model trained in a distinct source hospital.  Fine-tuning posits that deep models initialized using pre-trained models from a separate domain work better than randomly initialized models \cite{yosinski2014transferable}.  We train PHASE in a fine-tuning setting where upstream embedding models are trained in an OR target hospital initialized using the weights from the best model from the other OR hospital data set (detailed setup in Methods Section \ref{sec:meth:finetune}). 

\begin{figure}[!ht]
    \centering
    \includegraphics[width=.8\textwidth]{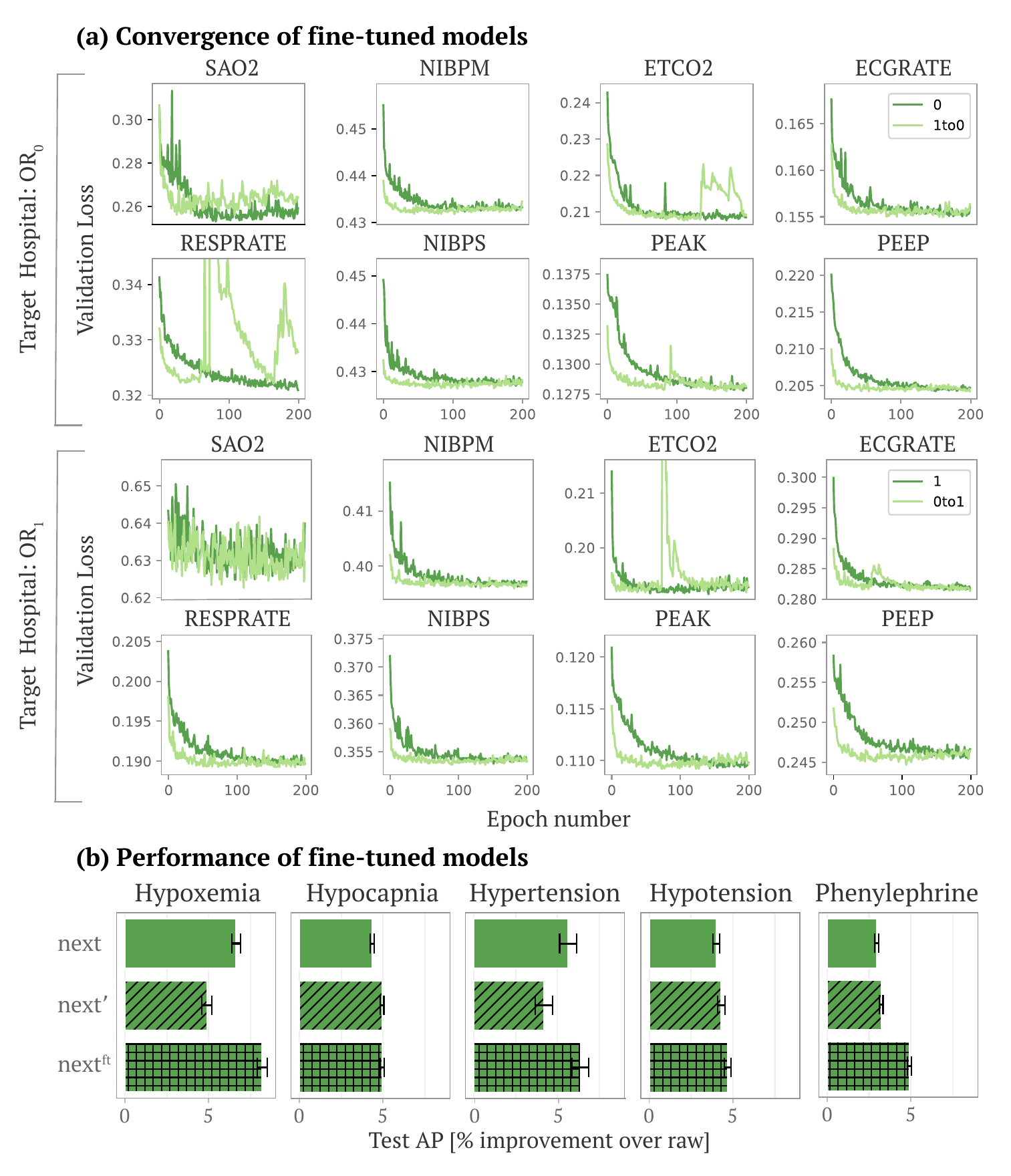}
    \caption{(a) The convergence of fine-tuned models.  The top eight plots fix OR$_0$ as the target dataset (we plot eight out of the total fifteen signals).  Dark green lines show the convergence of a randomly initialized LSTM trained in OR$_0$ and light green show the convergence of an LSTM trained in OR$_0$ initialized using weights from the best model in OR$_1$ (fine-tuning).  The bottom two rows show the analogous plots with OR$_1$ as the target dataset.   (b) The performance of GBT models trained on embeddings from standard embedding models (\textit{next}), transferred embedding models (\textit{next}'), and fine-tuned embedding models (\textit{next$^{\text{ft}}$}) (best models from light green in (a)).}
    \label{fig:ft_performance}
\end{figure}

We find that PHASE in the fine-tuned embedding setting boosts performance over both standard embedding (Section \ref{sec:results:performance}) and transferred embedding (Section \ref{sec:results:transference}) in Figure \ref{fig:ft_performance}b.  We focus on \textit{next} for the following experiment because it performed and generalized well across most outcomes in previous sections.  In Figure \ref{fig:ft_performance}, we evaluate the convergence and performance of fine-tuning LSTM embedding models.  
Figure \ref{fig:ft_performance}a shows the convergence of fine-tuned models.  The top two rows fix OR$_0$ as the target dataset.  Dark green lines show the convergence of a randomly initialized LSTM and light green show the convergence of an LSTM initialized using weights from the best model in OR$_1$.  The bottom two rows show the analogous plots with OR$_1$ as the target dataset.  
In Figure \ref{fig:ft_performance}a we see that fine-tuning LSTMs rather than training them from scratch consistently leads to much faster convergence.  In Figure \ref{fig:ft_performance}b, we see that LSTMs obtained from fine-tuning (\textit{next}$^{\text{ft}}$) consistently outperform those trained in a single dataset: standard embeddings (\textit{next}) and transferred embeddings (\textit{next}$'$).  These results indicate that end users can fine-tune PHASE LSTMs to boost performance at lower computational cost in comparison to training models from scratch.  Although fine-tuning is more computationally costly than a pre-trained model (transferred embedding), the performance gains from fine-tuning are more consistent.

\subsection{Validating models with local feature attributions}
\label{sec:results:interpretation}



\begin{figure}[!ht]
    \centering
    \includegraphics[width=\linewidth]{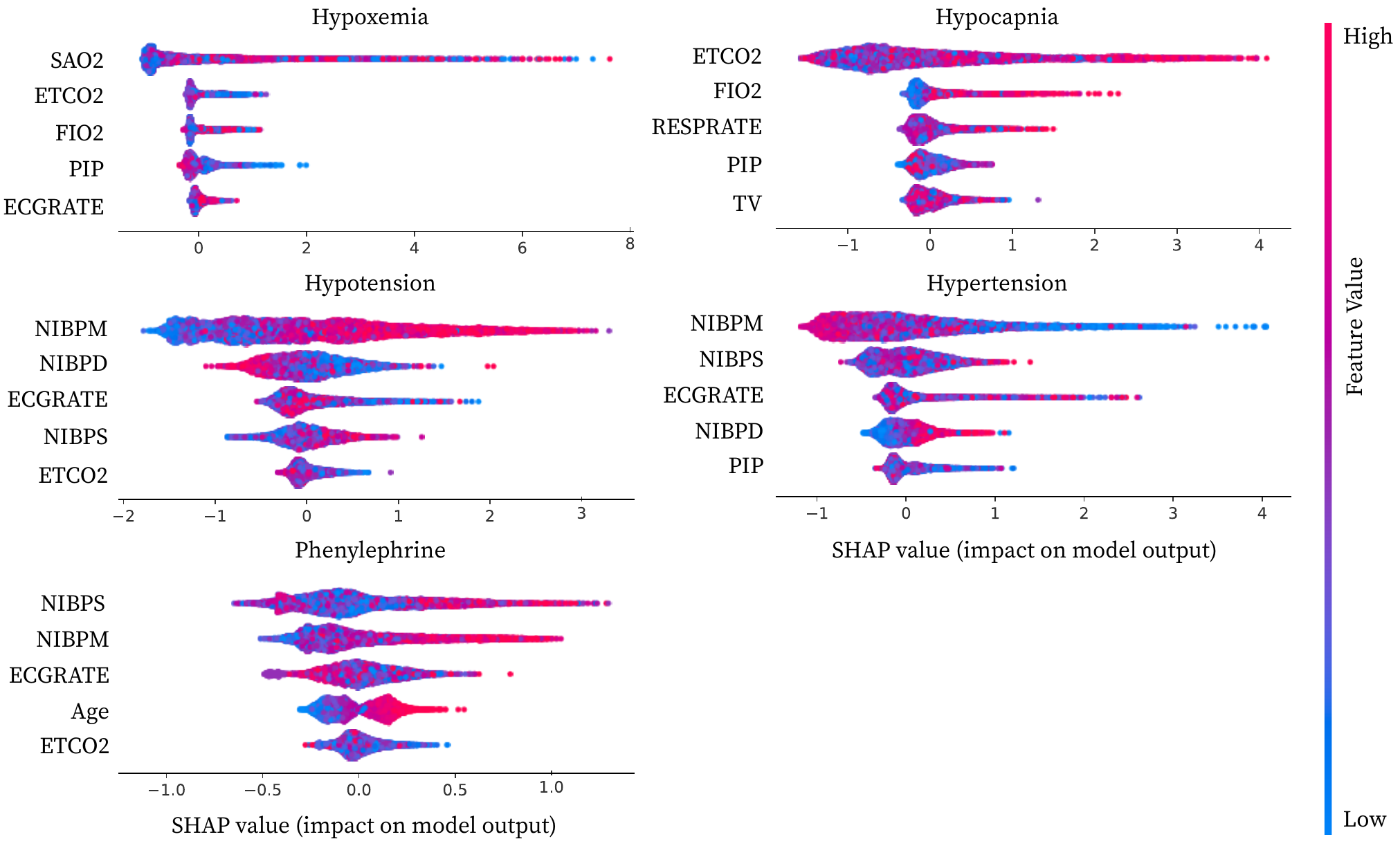}
    \caption{Local feature attribution summary plots for the top five most important variables from GBT models trained with \textit{next} embeddings in the target dataset OR$_0$.  In order to obtain attributions for each variable we explain each GBT using Interventional Tree Explainer.  This gives us attributions for \textit{next} embeddings for the fifteen physiological signal variables (200 dimensional embeddings for each) and six static variables.  We sum over embedding attributions to obtain the importance of a particular physiological signal variable.  On the x-axis we report this aggregated attribution value that indicates the variable’s cumulative impact on the model output.  The colors of the points are either the feature’s value for static variables or the sum over all \textit{next} embeddings for a given physiological signal variable. More detailed attributions in Supplementary Section \ref{sec:app:full_attr}.}
    \label{fig:interpretation}
\end{figure}


We summarize key variables used by downstream GBT models using summary plots (Figure \ref{fig:interpretation}).  In these plots, each point represents a feature's importance for a single sample, with the x-axis showing the feature's impact on the model's output and the colors indicates the feature's value (attribution method details in Methods Section \ref{sec:meth:shap}).  We focus on explaining GBT models trained on PHASE \textit{next} embeddings in terms of each variable because \textit{next} embeddings were performant across most of the outcomes we considered.  The colors are the sum of all features associated with a single signal variable (200 extracted features) which are not naturally interpretable because the embedding values can be arbitrarily positive or negative based on the embedding models.  

Standard approaches to train embedding models would use all signal variables as inputs to a single model.  These approaches are harder to interpret, because each embedding dimension may be dependent on multiple signals simultaneously.  Having per-signal embedding models as in PHASE allows us to clearly interpret each embedding as being dependent on a single physiological signal variable.

We validate important variables against prior literature for models trained on \textit{next} embeddings for all five outcomes (Figure \ref{fig:interpretation}).
For hypoxemia, the important variables includes variables logically connected to blood oxygen: \textit{SAO2}, \textit{ETCO2}, and \textit{FIO2} are all associated with the respiratory system, while and \textit{PIP} is tied to mechanical ventilation which is naturally linked to blood oxygen \cite{kiiski1992effect,dreyfuss1988high}.  
For hypocapnia \textit{ETCO2} is logically the most important feature.  Furthermore, using \textit{FIO2}, \textit{RESPRATE}, \textit{PIP}, and TV to forecast hypocapnia makes sense because these variables all relate to either ventilation or respiration.  As one would expect, for hypotension and hypertension, key variables are generally the three non-invasive blood pressure measurements: \textit{NIBPM}, \textit{NIBPD}, \textit{NIBPS}.  Furthermore, a number of studies validate the importance of \textit{ECGRATE} (heart rate measured from ECG signals) to forecasting hypotension and hypertension \cite{palatini2011role,morcet1999associations}.  Finally, phenelyphrine is typically administered during surgery in response to hypotension, thus validating the importance of \textit{NIBPS}, \textit{NIBPM}, and \textit{ECGRATE}.  Similarly, age being more important to forecast phenelyphrine use may be tied to its predictive relationship to hypotension as well as anesthesiologists' heightened vigilance to hypotension in the higher-risk older population \cite{lonjaret2014optimal}.





\section{Discussion}

This study explored machine learning techniques for forecasting adverse surgical outcomes.  Given the rates of adverse events in the operating room \cite{nilsson2016preventable,zegers2011incidence,kable2002adverse}, computational forecasting that provides advanced warning may be of widespread utility to medical practitioners.  This is especially the case given that the outcomes we considered (hypoxemia, hypocapnia, hypotension, and hypertension) are all tied to a number of harmful physiological effects.

This work also shows physiological signal embeddings are effective in several settings.  We demonstrate that standard embedding using LSTMs improves the performance of downstream models (GBT and MLP), which implies that pipelines utilizing deep networks to embed physiological signals are effective for electronic healthcare record data.   Next, we show that PHASE embedding models work almost equally well in a transferred embedding setting as in a standard embedding setting, and, in fact, work better than randomly initialized models if fine-tuned.  This implies that sharing pre-trained networks can improve downstream models in terms of computational needs and predictive performance.

PHASE uses independently trained LSTMs for each signal variable.  Surprisingly, we demonstrate that our per-signal approach outperforms a jointly trained embedding model LSTM (see Supplementary Section \ref{sec:app:joint}).  Furthermore, having each LSTM associated with a single physiological signal actually proves to be an advantage.  Hospitals often collect different sets of physiological signal variables; to address this heterogeneity, target hospitals with different but overlapping variables to a source hospital can use the embedding models for the variables which they both have (see Supplementary Section \ref{sec:app:heterogeneous}).  

One limitation of PHASE is that although sharing models reveals less information than sharing data, it is possible to use model inversion attacks on the PHASE embedding models \cite{fredrikson2015model} to find physiological signals similar to the training data.  Although we attempted to use differentially private versions of stochastic gradient descent \cite{abadi2016deep} to train our embedding models, the randomness inserted in the training process made it difficult to train effective models.  We leave investigation and development of effective privacy preserving techniques to train such models to future work.  Another limitation of our data is that the embedding models only apply to physiological signals sampled once per minute.  We leave exploration of adapting models to accommodate multiple sampling frequencies to future work as well.  Finally, it should be said that there is complementary work discussing deep learning for electrocardiograms \cite{salem2018ecg,mathews2018novel} and electroencephalograms \cite{oh2018deep}.  We focus primarily on minute by minute physiological signals collected within an operating room setting.  As such, although we do have an ECGRATE variable, we do not directly use the electrocardiogram signals.

Our work takes an important step forward in applying machine learning to the domain of physiological signals.  Our approach differs from previous studies, which did not explore physiological signal transfer learning across multiple hospitals or analyze self-supervised approaches for training deep models.

Drawing on parallels from computer vision (CV) and natural language processing (NLP), both exemplars of transfer learning, physiological signals are well suited to neural network embeddings (i.e., transformations of original inputs into a space better suited to make predictions).  In particular, CV and NLP share two notable traits with physiological signals. The first is \textit{consistency}. The CV domain has consistent features:  edges, colors, and other visual attributes \cite{raina2007self,shin2016deep}; the NLP domain uses a particular language with semantic relationships consistent across bodies of text \cite{conneau2017supervised}. For sequential signals, we saw that physiological patterns are consistent, because PHASE generalized across hospitals in a transferred embedding setting. The second attribute is \textit{complexity}. Each of these domains is sufficiently complex to make learning embeddings non-trivial. These factors suggest that individual research scientists must make redundant efforts to learn embeddings that may ultimately be very similar. To avoid this problem, NLP and CV have made significant progress on standardizing and evaluating their embeddings; in the health domain, physiological signals are a natural next step.  More significantly, the use of physiological signals is constrained by patient privacy; this makes it difficult to share \textit{data} between hospitals. However, sharing \textit{models} between hospitals does not directly expose patient information.  Sharing models in this way could allow machine learning for physiological signals to see similarly large advances as in computer vision and natural language.

\section{Methods}

\subsection{Datasets}
\label{sec:data_cohort}

The operating room (OR) datasets were collected via the Anesthesia Information Management System (AIMS), which includes static information as well as real-time measurements of physiological signals sampled minute by minute.   OR$_0$ was drawn from an academic medical center and OR$_1$ from a trauma center.  Two marked differences between the patient distributions of OR$_0$ and OR$_1$ are the gender ratio (57\% females in the academic medical center versus 38\% in the trauma center) and the proportion of ASA codes that are classified as emergencies (7.65\% emergencies versus 15.31\%). ICU$_\text{M}$ is a sub-sampled version drawn from PhysioNet's publicly available MIMIC dataset, which contains data obtained from an intensive care unit (ICU) in Boston, Massachusetts \cite{johnson2016}.  Although ICU$_\text{M}$ data contains several physiological signals sampled at a high frequency, we solely used a minute-by-minute \textit{SAO2} signal for our experiments because other physiological signals had a substantial amount of missingness.  Furthermore, ICU$_\text{M}$ contained neonatal data that we filtered out.  For all three datasets, any remaining missing values in the signal features were imputed by the mean, and each feature was standardized to have unit mean and variance for training neural networks.  Additional details about the distributions of patients in all three datasets are shown in Table \ref{tab:hospitals}, and a list of prevalent diagnoses is presented in Supplementary Section \ref{sec:app:diagnoses}.

\subsection{Set-up}
\label{sec:meth:features}

For our datasets, we considered a distribution of hospital stays $\mathcal{P}$.  Since we wanted to forecast an adverse event in time, we defined samples by first drawing a hospital stay $P\sim \mathcal{P}$ and then drawing a time point $t\sim (1,\cdots, len(P))$.  For the rest of this set-up, we assume we are operating with samples $i$ defined by $t,P$.

\subsubsection{Variables}
\label{sec:meth:variables}

Many variables are associated with each hospital stay.  We distinguished between static variables (that are constant throughout the course of a patient's stay and are solely determined by $P$) and dynamic variables (that change over time and are determined by $P$ and $t$).  We partition each sample $i$'s ($i$ is implicitly determined by $P$ and $t$) variables into two distinct sets:
\begin{align}
X^i=(\underbrace{X^i_{s_1},\cdots,X^i_{s_6}}_{\text{Static variables}},\underbrace{X^i_{d_1},\cdots,X^i_{d_{15}}}_{\text{Dynamic variables}})
\end{align}

The six static variables $(X^i_{s_1},\cdots,X^i_{s_6})$ that do not change over the course of a surgery are: {\it Height}, {\it Weight}, {\it ASA Code}, {\it ASA Code Emergency}, {\it Gender}, and {\it Age}. 

Furthermore, we utilized fifteen physiological signals for our dynamic variables ($X^i_{d_1},\cdots,X^i_{d_{15}}$): 
\begin{itemize}
\itemsep0em 
\item {\it SAO2} - Blood oxygen saturation
\item {\it ETCO2} - End-tidal carbon dioxide
\item {\it NIBP[S/M/D]} - Non-invasive blood pressure (systolic, mean, diastolic)
\item {\it FIO2} - Fraction of inspired oxygen
\item {\it ETSEV/ETSEVO} - End-tidal sevoflurane
\item {\it ECGRATE} - Heart rate from ECG
\item {\it PEAK} - Peak ventilator pressure
\item {\it PEEP} - Positive end-expiratory pressure
\item {\it PIP} - Peak inspiratory pressure
\item {\it RESPRATE} - Respiration rate
\item {\it TEMP1} - Body temperature
\item {\it PHENYL} - Whether phenylephrine was administered.  We only use this as an output variable and not as an input.
\end{itemize}

To index the dynamic variables we used the following notation to denote minutes $a$ to $b$ (where $b>a$) of a particular signal:
\begin{align}
X_{d_j}^i[a:b]\in \mathbb{R}^{b-a}
\end{align}

\subsubsection{Outcomes} 

We focused on binary outcomes (i.e., downstream prediction tasks):
\begin{align}
y^i\in \{0,1\} 
\end{align}

Our adverse events define the outcome as a function ($g(\cdot), e.g., g(\cdot)=min(\cdot)<C$) of the next five minutes of a physiological signal ($X^i_{d_j}$):
\begin{align}
y^i=g(X^i_{d_j}[t+1:t+5])   
\end{align}

Specifically, we focused on health forecasting tasks; forecasting tasks facilitate preventive healthcare by helping healthcare providers mitigate risk preemptively \cite{soyiri2013overview}.  In particular, we considered the following five tasks (which all focus on the next five minutes of surgery): 
\begin{itemize}
\itemsep0em 
\item \textit{Hypoxemia:} was blood oxygen less than 93?
\begin{align}
\min(X^i_\textit{SAO2}[t+1:t+5])< 93
\end{align}
\item \textit{Hypocapnia:} was end tidal carbon dioxide less than 35?
\begin{align}
\min (X^i_\textit{ETCO2}[t+1:t+5])<35
\end{align}
\item \textit{Hypotension:} was mean blood pressure less than 60?
\begin{align}
\min (X^i_\textit{NIBPM}[t+1:t+5])<60
\end{align}
\item \textit{Hypertension:} was mean blood pressure higher than 110?
\begin{align}
\min (X^i_\textit{NIBPM}[t+1:t+5])>110
\end{align}
\item \textit{Phenylephrine:} was phenylephrine administered?
\begin{align}
\max(X^i_\textit{PHENYL}[t+1:t+5]) = 1
\end{align}
\end{itemize}
More details about our labelling schemes are in Supplementary Section \ref{sec:app:labelling}.

\subsection{Embeddings (i.e. features)}
We define variables (e.g., height, blood oxygen, etc.) separately from embeddings (e.g., height, minute 20 of blood oxygen, etc.) which the downstream prediction models are trained on.  Notationally, we denote embeddings as lower case:
\begin{equation*}
x^i=(x^i_{s_1},\cdots,x^i_{s_6},x^i_{d_1},\cdots,x^i_{d_{15}}).
\end{equation*}

We embed the dynamic variables, with a function $U_{d_k;E}$ of the past sixty minutes of the physiological signal variable: 
\begin{equation*}
    x^i_{d_k}=U_{d_k;E}(X^i_{d_k}[t-59:t]), \forall k\in 1,\cdots,15, E\in \{raw,ema,rand,auto,next,min,hypo\}.
\end{equation*}

We use the static variables as is: $x^i_{s_k}=X^i_{s_k}, \forall k\in 1,\cdots,6$.  For GBT downstream models we do not transform the static variables; however, for the LSTM downstream models we do normalize them.  Unlike dynamic variables, extracting features from the static variables does not significantly improve performance of downstream models.

\subsection{Downstream prediction model}

The downstream prediction models $D$ are used to evaluate different types of embeddings.  They are trained on the embedded samples $x^i$ drawn from a target hospital $H_t$.  $D$ minimizes binary cross entropy loss to forecast adverse outcomes $y^i$ defined as a function of the future five minutes of a physiological signal (for example hypoxemia would be $\min (X^i_{d_{SAO2}}[t+1:t+5])<93$, where $X^i_{d_{SAO2}}[t+1:t+5]$ denotes the future five minutes of the blood oxygen variable for sample $i$).

\subsection{Dynamic embedding}

For dynamic variables, we made two important decisions.  The first was how much of the signal to use.  To make fair comparisons, we gave all models access only to the 60 minutes of the signal prior to the outcome (which starts at $t+1$):
\begin{align}
X^i_{d_j}[t-59:t]
\end{align}

The second important decision was how to embed a signal ($X^i_{d_j}$).  Two natural embeddings are: (1) to use the sixty minutes as is (\textit{raw}):
\begin{align}
x^i_{d_j} = X^i_{d_j}[t-59:t] \in \mathbb{R}^{60}
\end{align}
Where $U_{d_j;raw}$ is the identify function.

and (2) to use exponential moving averages and variances as the embedding function $U_{d_j;ema}$ (\textit{ema}) \cite{lundberg2018explainable}:
\begin{align}
x^i_{d_j} = (EMA(X^i_{d_j}[t-59:t],\alpha=0.1),EMA(X^i_{d_j}[t-59:t],\alpha=1),\\EMA(X^i_{d_j}[t-59:t],\alpha=5),EMV(X^i_{d_j}[t-59:t],\alpha=5)) \in \mathbb{R}^{4}
\end{align}

where the exponential moving average is defined as:
\begin{align}
EMA_\tau &= \alpha \times X^i_{d_j}[\tau] + (1-\alpha)\times EMA_{\tau-1}, \forall \tau>t-59\\
EMA_{t-59} &= X^i_{d_j}[t-59]\\
EMA(X^i_{d_j}[t-59:t],\alpha) &= EMA_t
\end{align}

and the exponential moving variance is defined as:
\begin{align}
\delta_\tau&=X^i_{d_j}[\tau]-EMA_{\tau-1}\\
EMA_\tau&=EMA_{\tau-1}+\alpha \times \delta_\tau\\
EMV_\tau&=(1-\alpha)\times (EMV_{\tau-1}+\alpha\times \delta_\tau^2)\\
EMV(X^i_{d_j}[t-59:t],\alpha=5)&=EMV_t
\end{align}

\subsubsection{LSTM embedding}
\label{sec:meth:nn_feature_extraction}

To better extract features from (embed) each physiological signal variable ($X^i_{d_j}$), we utilized per-signal neural networks (LSTMs) trained in a source hospital $H_s$.  The LSTMs $L^{H_s}_{d_j;E}$ are trained for each physiological signal (we show that per-signal embedding models worked better than a single LSTM trained on all signals jointly in Supplementary Section \ref{sec:app:joint}) to minimize a loss function (dependent on the embedding type $E$) with the past sixty minutes of signal $d_k$ as the input:
\begin{equation*}
\mathcal{L}_{E}(L^{H_s}_{d_k;E}(X^i_{d_k}[t-59:t]), y^i_{E})
\end{equation*}

\begin{table}[!ht]
\centering
\begin{tabular}{|l|l|l|l|} \hline
     $E$ & Domain                                      & Range (Upstream Task) & $\mathcal{L}_E$                                          \\ \hline
\textit{rand} & $X^i_{d_j}[t-59:t] \in \mathbb{R}^{60}$ & $\emptyset$ & $\emptyset$                                  \\
\textit{auto} & $X^i_{d_j}[t-59:t] \in \mathbb{R}^{60}$ & $X^i_{d_j}[t-59:t] \in \mathbb{R}^{60}$ & MSE       \\
\textit{next} & $X^i_{d_j}[t-59:t] \in \mathbb{R}^{60}$ & $X^i_{d_j}[t+1:t+5] \in \mathbb{R}^5$ & MSE            \\
\textit{min}  & $X^i_{d_j}[t-59:t] \in \mathbb{R}^{60}$ & $\min(X^i_{d_j}[t+1:t+5]) \in \mathbb{R}^1$ & MSE      \\
\textit{hypo} & $X^i_{d_j}[t-59:t] \in \mathbb{R}^{60}$ & $y^i \in \{0,1\}$ & BCE \\ \hline
\end{tabular}
\caption{Inputs and outputs for our per-signal upstream LSTMs.}
\label{tab:self_superviseH_tasks}
\end{table}

Table \ref{tab:self_superviseH_tasks} describes the different tasks we used to train LSTMs upstream embedding models including the three self-supervised labels (\textit{next}, \textit{min}, \textit{hypo}) we proposed in PHASE.  More specifically, $U_{d_j;E}=h\circ L^{H_s}_{d_j;E}$, where the composition $h\circ L$ signifies removing the output layer of $L$ to obtain a function that maps the past sixty minutes of $d_k$ to the activations of the final hidden layer in $L$.  For the \textit{rand} embedding the models $L_{d_k;rand}$ are LSTM models with random weights.  There is no source hospital, because the models are not trained.  Then, \textit{auto}, \textit{next}, and \textit{min} embeddings set $\mathcal{L}_E$ to mean squared error.  However, the outcomes differ for each: $y^i_{auto}=X^i_{d_k}[t-59:t], y^i_{next}=X^i_{d_k}[t+1:t+5], y^i_{mind}=\min(X^i_{d_k}[t+1:t+5])$ (note that these outcomes are self-supervised).  Finally, \textit{hypo} embeddings set $\mathcal{L}_E$ to binary cross entropy loss and the outcome is set to be the same as the downstream task $y^i$.  Since several of our downstream outcomes were tied to too-low (``hypo'') signals, the approaches in Table \ref{tab:self_superviseH_tasks} were ordered by distance to the downstream task.


We used the following notation to denote an LSTM trained to convergence using $X^i_{d_j}$ drawn from the source hospital dataset $H_s$ using inputs and outputs specified by the task in Table \ref{tab:self_superviseH_tasks}:
\begin{align}
L^{H_s}_{d_j;\textit{task}}
\end{align}

As an example, $L_{d_j;next}^{\text{OR}_0}$ indicates that the LSTM was trained for signal $X_{d_j}^i$ with inputs $X_{d_j}^i[t-59:t]$ and outputs $X_{d_j}^i[t+1:t+5]$ on data drawn from $\text{OR}_0$.

To describe the features associated with the neural network embedding approaches, we removed the output layer of the network and embedded each signal using the final hidden layer of the network.  We denote this as:
\begin{align}
x^i_{d_j}\equiv h\circ L_{d_j;next}^{H_s}(X_{d_j}^i[t-59:t]) \in \mathbb{R}^{200}
\end{align}
where $h$ removes the output layer of network $L$ and 200 is the number of hidden nodes in $L$.

As an example, if our target dataset was OR$_0$, then our physiological signal features for \textit{next} would be:
\begin{align}
x^i_{d_j}\equiv h\circ L_{d_j;next}^{\text{OR}_0}(X_{d_j}^i[t-59:t]) \in \mathbb{R}^{200}
\end{align}

\subsubsection{Transferred embedding}
\label{sec:meth:transferlearning}


To evaluate transfer learning, we denoted a target hospital dataset $H_t$ (the domain in which we trained the downstream prediction model on embedded variables) and a source hospital dataset $H_s$ (the domain in which we trained our upstream embedding models).  In the transference experiments (denoted used superscripts next to the embedding type $E$: \textit{task}$'$ and \textit{task}$^M$) we train our upstream embedding models in a source hospital that is different to the target hospital ($H_s\neq H_t$).

By default, without the superscript, the source domain matched the target domain ($H_s=H_t$).  With an apostrophe, the source domain was the remaining operating room dataset ($H_s=\text{OR}_0$ if $H_t=\text{OR}_1$ or $H_s=\text{OR}_1$ if $H_t=\text{OR}_0$).  As an example, if our target dataset was OR$_0$, then our physiological signal features for \textit{next}$'$ would be:
\begin{align}
x^i_{d_j}\equiv h\circ L_{d_j;next}^{\text{OR}_1}(X_{d_j}^i[t-59:t]) \in \mathbb{R}^{200}
\end{align}

Finally, for \textit{task}$^M$, the source domain for the LSTM embedding model for SAO2 was ICU$_\text{M}$ ($H_s=\text{ICU}_\text{M}$), and the remaining models were trained in a source domain that matched the target domain ($H_s=H_t$).  As an example, if our target dataset was OR$_0$, then our physiological signal features for \textit{next}$'$ would be:
\begin{align}
x^i_{d_j}\equiv h\circ L_{d_j;next}^{\text{ICU}_\text{M}}(X_{d_j}^i[t-59:t]) &\in \mathbb{R}^{200}\text{ for SAO2}\\
x^i_{d_j}\equiv h\circ L_{d_j;next}^{\text{OR}_0}(X_{d_j}^i[t-59:t]) &\in \mathbb{R}^{200}\text{ for all other signals}
\end{align}

\subsubsection{Fine-tuned embedding}
\label{sec:meth:finetune}

The fine-tuning approach (denoted as next$^{\text{ft}}$) considers fine tuning models between operating room datasets.  If we assume a fixed target dataset $H_t=$OR$_0$.  Then, as before, we denote an LSTM trained to convergence on data from OR$_1$ to be:
\begin{align}
L^{\text{OR}_1}_{d_j;next}
\end{align}

For fine-tuning, we used the LSTM trained on samples drawn from OR$_1$ (which crucially was not the same as the target dataset) to initialize an LSTM which we then trained until convergence on samples drawn from OR$_0$.  Notationally, we describe this as:
\begin{align}
L^{\text{OR}_1\to \text{OR}_0}_{d_j;next}
\end{align}

The features for dynamic variables under the fine-tuning approach for $H_t=\text{OR}_0$ were:
\begin{align}
x^i_{d_j}\equiv h\circ L_{d_j;next}^{\text{OR}_1\to \text{OR}_0}(X_{d_j}^i[t-59:t]) \in \mathbb{R}^{200}
\end{align}

\subsubsection{Jointly Trained Upstream Model}
\label{sec:meth:joint}

The jointly trained upstream model (denoted as \textit{next}$_\text{m}$) involved training an LSTM for several signals simultaneously.  To do so, we optimized an LSTM for forecasting the next five minutes of all our physiological signals, which we denote as:
\begin{align}
L^{H_s}_{d_1,\cdots,d_{15};next}
\end{align}

Then, the features for dynamic variables under the jointly trained multi-signal model were:
\begin{align}
x^i_{d_1},\cdots,x^i_{d_{15}}=h\circ L^{H_s}_{d_1,\cdots,d_{15};next}(X_{d_1}^i[t-59:t],\cdots,X_{d_{15}}^i[t-59:t])
\end{align}

\subsubsection{Local Feature Attributions}
\label{sec:meth:shap}

To obtain explanations, we utilized Interventional Tree Explainer, which provides exact Shapley values with an interventional conditional expectation set function (feature attributions with game-theoretic properties) for complex tree-based models \cite{lundberg2020local,lundberg2017unified}.  The Shapley values serve as local feature attributions $\phi(f,x^i)$ that indicate how much each feature in $x^i$ contributed to a single downstream prediction $D(x^i)$.  Positive attribution means that the feature generally increases the output of the model (risk of adverse events) and negative attribution means that the feature generally decreases the output.  Shapley values have been used to explain models in a wide variety of applications including biology \cite{kim2020predicting}, medicine \cite{penny2020machine}, finance \cite{theil2020predicting}, and more.

\newpage

\printbibliography

@article{nilsson2016preventable,
  title={Preventable adverse events in surgical care in Sweden: a nationwide review of patient notes},
  author={Nilsson, Lena and Risberg, Madeleine Borgstedt and Montgomery, Agneta and Sj{\"o}dahl, Rune and Schildmeijer, Kristina and Rutberg, Hans},
  journal={Medicine},
  volume={95},
  number={11},
  year={2016},
  publisher={Wolters Kluwer Health}
}

@article{zegers2011incidence,
  title={The incidence, root-causes, and outcomes of adverse events in surgical units: implication for potential prevention strategies},
  author={Zegers, Marieke and de Bruijne, Martine C and de Keizer, Bertus and Merten, Hanneke and Groenewegen, Peter P and van der Wal, Gerrit and Wagner, Cordula},
  journal={Patient safety in surgery},
  volume={5},
  number={1},
  pages={13},
  year={2011},
  publisher={Springer}
}

@article{kable2002adverse,
  title={Adverse events in surgical patients in Australia},
  author={Kable, AK and Gibberd, RW and Spigelman, AD},
  journal={International Journal for Quality in Health Care},
  volume={14},
  number={4},
  pages={269--276},
  year={2002},
  publisher={Oxford University Press}
}

@article{weiser2016size,
  title={Size and distribution of the global volume of surgery in 2012},
  author={Weiser, Thomas G and Haynes, Alex B and Molina, George and Lipsitz, Stuart R and Esquivel, Micaela M and Uribe-Leitz, Tarsicio and Fu, Rui and Azad, Tej and Chao, Tiffany E and Berry, William R and others},
  journal={Bull World Health Organ},
  volume={94},
  number={3},
  pages={201--209F},
  year={2016}
}

@article{lundberg2018explainable,
  title={Explainable machine-learning predictions for the prevention of hypoxaemia during surgery},
  author={Lundberg, Scott M and Nair, Bala and Vavilala, Monica S and Horibe, Mayumi and Eisses, Michael J and Adams, Trevor and Liston, David E and Low, Daniel King-Wai and Newman, Shu-Fang and Kim, Jerry and others},
  journal={Nature biomedical engineering},
  volume={2},
  number={10},
  pages={749--760},
  year={2018},
  publisher={Nature Publishing Group}
}

@article{steiner2017,
  title={Surgeries in Hospital-Based Ambulatory Surgery and Hospital Inpatient Settings, 2014},
  author={Claudia A. Steiner and Zeynal Karaca and Brian J.
Moore and Melina C. Imshaug and Gary Pickens},
  journal={HCUP Statistical Brief},
  year={2017},
  publisher={AHRQ}
}

@article{wen2016,
  title={An All-Payer View of Hospital Discharge to Postacute Care, 2013},
  author={Tian Wen},
  journal={HCUP Statistical Brief},
  year={2016}
}

@ARTICLE{7426826, 
author={N. Tajbakhsh and J. Y. Shin and S. R. Gurudu and R. T. Hurst and C. B. Kendall and M. B. Gotway and J. Liang}, 
journal={IEEE Transactions on Medical Imaging}, 
title={Convolutional Neural Networks for Medical Image Analysis: Full Training or Fine Tuning?}, 
year={2016}, 
volume={35}, 
number={5}, 
pages={1299-1312}, 
doi={10.1109/TMI.2016.2535302}, 
ISSN={0278-0062}, 
month={5},}

@incollection{ravishankar2016understanding,
  title={Understanding the mechanisms of deep transfer learning for medical images},
  author={Ravishankar, Hariharan and Sudhakar, Prasad and Venkataramani, Rahul and Thiruvenkadam, Sheshadri and Annangi, Pavan and Babu, Narayanan and Vaidya, Vivek},
  booktitle={Deep Learning and Data Labeling for Medical Applications},
  pages={188--196},
  year={2016},
  publisher={Springer}
}

@article{LV201455,
title = "Transfer learning based clinical concept extraction on data from multiple sources",
journal = "Journal of Biomedical Informatics",
volume = "52",
pages = "55 - 64",
year = "2014",
note = "Special Section: Methods in Clinical Research Informatics",
issn = "1532-0464",
doi = "https://doi.org/10.1016/j.jbi.2014.05.006",
url = "http://www.sciencedirect.com/science/article/pii/S1532046414001233",
author = "Xinbo Lv and Yi Guan and Benyang Deng"
}

@article{majumder2017wearable,
  title={Wearable sensors for remote health monitoring},
  author={Majumder, Sumit and Mondal, Tapas and Deen, M},
  journal={Sensors},
  volume={17},
  number={1},
  pages={130},
  year={2017},
  publisher={Multidisciplinary Digital Publishing Institute}
}

@article{roski2014creating,
  title={Creating value in health care through big data: opportunities and policy implications},
  author={Roski, Joachim and Bo-Linn, George W and Andrews, Timothy A},
  journal={Health affairs},
  volume={33},
  number={7},
  pages={1115--1122},
  year={2014}
}

@article{orphanidou2019review,
  title={A review of big data applications of physiological signal data},
  author={Orphanidou, Christina},
  journal={Biophysical reviews},
  volume={11},
  number={1},
  pages={83--87},
  year={2019},
  publisher={Springer}
}

@article{kohli2017medical,
  title={Medical image data and datasets in the era of machine learning—whitepaper from the 2016 C-MIMI meeting dataset session},
  author={Kohli, Marc D and Summers, Ronald M and Geis, J Raymond},
  journal={Journal of digital imaging},
  volume={30},
  number={4},
  pages={392--399},
  year={2017},
  publisher={Springer}
}

@article{fahimi2018inter,
  title={Inter-subject transfer learning with end-to-end deep convolutional neural network for EEG-based BCI},
  author={Fahimi, Fatemeh and Zhang, Zhuo and Goh, Wooi Boon and Lee, Tih-Shih and Ang, Kai Keng and Guan, Cuntai},
  journal={Journal of neural engineering},
  year={2018},
  publisher={IOP Publishing}
}

@article{gupta2019transfer,
  title={Transfer Learning for Clinical Time Series Analysis using Deep Neural Networks},
  author={Gupta, Priyanka and Malhotra, Pankaj and Narwariya, Jyoti and Vig, Lovekesh and Shroff, Gautam},
  journal={arXiv preprint arXiv:1904.00655},
  year={2019}
}

@article{johnson2016, title={MIMIC-III, a freely accessible critical care database}, volume={3}, DOI={10.1038/sdata.2016.35}, journal={Scientific Data}, author={Johnson, Alistair E.w. and Pollard, Tom J. and Shen, Lu and Lehman, Li-Wei H. and Feng, Mengling and Ghassemi, Mohammad and Moody, Benjamin and Szolovits, Peter and Celi, Leo Anthony and Mark, Roger G. and et al.}, year={2016}, pages={160035}}

@inproceedings{raina2007self,
  title={Self-taught learning: transfer learning from unlabeled data},
  author={Raina, Rajat and Battle, Alexis and Lee, Honglak and Packer, Benjamin and Ng, Andrew Y},
  booktitle={Proceedings of the 24th international conference on Machine learning},
  pages={759--766},
  year={2007},
  organization={ACM}
}

@article{shin2016deep,
  title={Deep convolutional neural networks for computer-aided detection: CNN architectures, dataset characteristics and transfer learning},
  author={Shin, Hoo-Chang and Roth, Holger R and Gao, Mingchen and Lu, Le and Xu, Ziyue and Nogues, Isabella and Yao, Jianhua and Mollura, Daniel and Summers, Ronald M},
  journal={IEEE transactions on medical imaging},
  volume={35},
  number={5},
  pages={1285--1298},
  year={2016},
  publisher={IEEE}
}

@article{conneau2017supervised,
  title={Supervised learning of universal sentence representations from natural language inference data},
  author={Conneau, Alexis and Kiela, Douwe and Schwenk, Holger and Barrault, Loic and Bordes, Antoine},
  journal={arXiv preprint arXiv:1705.02364},
  year={2017}
}

@article{korner1959circulatory,
  title={Circulatory adaptations in hypoxia},
  author={Korner, PI},
  journal={Physiological reviews},
  volume={39},
  number={4},
  pages={687--730},
  year={1959}
}

@article{ehrenfeld2010incidence,
  title={The incidence of hypoxemia during surgery: evidence from two institutions},
  author={Ehrenfeld, Jesse M and Funk, Luke M and Van Schalkwyk, Johan and Merry, Alan F and Sandberg, Warren S and Gawande, Atul},
  journal={Canadian Journal of Anesthesia/Journal canadien d'anesth{\'e}sie},
  volume={57},
  number={10},
  pages={888--897},
  year={2010},
  publisher={Springer}
}

@article{pollard1977some,
  title={Some adverse physiological effects of hypocarbia and methods of maintaining normocarbia during controlled ventilation—a review},
  author={Pollard, Brian and Gibb, David B},
  journal={Anaesthesia and intensive care},
  volume={5},
  number={2},
  pages={113--121},
  year={1977},
  publisher={SAGE Publications Sage UK: London, England}
}

@article{lienhart2006survey,
  title={Survey of anesthesia-related mortality in France},
  author={Lienhart, Andre and Auroy, Yves and Pequignot, Francoise and Benhamou, Dan and Warszawski, Josiane and Bovet, Martine and Jougla, Eric},
  journal={Anesthesiology: The Journal of the American Society of Anesthesiologists},
  volume={105},
  number={6},
  pages={1087--1097},
  year={2006},
  publisher={The American Society of Anesthesiologists}
}

@article{chang2000adverse,
  title={Adverse effects of limited hypotensive anesthesia on the outcome of patients with subarachnoid hemorrhage},
  author={Chang, Han Soo and Hongo, Kazuhiro and Nakagawa, Hiroshi},
  journal={Journal of neurosurgery},
  volume={92},
  number={6},
  pages={971--975},
  year={2000},
  publisher={Journal of Neurosurgery Publishing Group}
}

@article{basali2000relation,
  title={Relation between perioperative hypertension and intracranial hemorrhage after craniotomy},
  author={Basali, Ayman and Mascha, Edward J and Kalfas, Iain and Schubert, Armin},
  journal={Anesthesiology: The Journal of the American Society of Anesthesiologists},
  volume={93},
  number={1},
  pages={48--54},
  year={2000},
  publisher={The American Society of Anesthesiologists}
}

@article{varon2008perioperative,
  title={Perioperative hypertension management},
  author={Varon, Joseph and Marik, Paul E},
  journal={Vascular health and risk management},
  volume={4},
  number={3},
  pages={615},
  year={2008},
  publisher={Dove Press}
}

@article{kee2004prophylactic,
  title={Prophylactic phenylephrine infusion for preventing hypotension during spinal anesthesia for cesarean delivery},
  author={Kee, Warwick D Ngan and Khaw, Kim S and Ng, Floria F and Lee, Bee B},
  journal={Anesthesia \& Analgesia},
  volume={98},
  number={3},
  pages={815--821},
  year={2004},
  publisher={LWW}
}

@article{soyiri2013overview,
  title={An overview of health forecasting},
  author={Soyiri, Ireneous N and Reidpath, Daniel D},
  journal={Environmental health and preventive medicine},
  volume={18},
  number={1},
  pages={1},
  year={2013},
  publisher={BioMed Central}
}

@article{jeremitsky2003harbingers,
  title={Harbingers of poor outcome the day after severe brain injury: hypothermia, hypoxia, and hypoperfusion},
  author={Jeremitsky, Elan and Omert, Laurel and Dunham, C Michael and Protetch, Jack and Rodriguez, Aurelio},
  journal={Journal of Trauma and Acute Care Surgery},
  volume={54},
  number={2},
  pages={312--319},
  year={2003},
  publisher={LWW}
}

@article{doshi2017towards,
  title={Towards a rigorous science of interpretable machine learning},
  author={Doshi-Velez, Finale and Kim, Been},
  journal={arXiv preprint arXiv:1702.08608},
  year={2017}
}

@inproceedings{lundberg2017unified,
  title={A unified approach to interpreting model predictions},
  author={Lundberg, Scott M and Lee, Su-In},
  booktitle={Advances in Neural Information Processing Systems},
  pages={4765--4774},
  year={2017}
}

@article{kiiski1992effect,
  title={Effect of tidal volume on gas exchange and oxygen transport in the adult respiratory distress syndrome},
  author={Kiiski, Ritva and Takala, Jukka and Kari, Aarno and Milic-Emili, J},
  journal={American Review of Respiratory Disease},
  volume={146},
  pages={1131--1131},
  year={1992},
  publisher={American Lung Association}
}

@article{dreyfuss1988high,
  title={High inflation pressure pulmonary edema: respective effects of high airway pressure, high tidal volume, and positive end-expiratory pressure},
  author={Dreyfuss, Didier and Soler, Paul and Basset, Guy and Saumon, Georges},
  journal={American Review of Respiratory Disease},
  volume={137},
  number={5},
  pages={1159--1164},
  year={1988},
  publisher={American Lung Association}
}

@article{palatini2011role,
  title={Role of elevated heart rate in the development of cardiovascular disease in hypertension},
  author={Palatini, Paolo},
  journal={Hypertension},
  volume={58},
  number={5},
  pages={745--750},
  year={2011},
  publisher={Am Heart Assoc}
}

@article{morcet1999associations,
  title={Associations between heart rate and other risk factors in a large French population},
  author={Morcet, Jean-Fran{\c{c}}ois and Safar, Michel and Thomas, Fr{\'e}d{\'e}rique and Guize, Louis and Benetos, Athanase},
  journal={Journal of hypertension},
  volume={17},
  number={12},
  pages={1671--1676},
  year={1999},
  publisher={LWW}
}

@article{lonjaret2014optimal,
  title={Optimal perioperative management of arterial blood pressure},
  author={Lonjaret, Laurent and Lairez, Olivier and Minville, Vincent and Geeraerts, Thomas},
  journal={Integrated blood pressure control},
  volume={7},
  pages={49},
  year={2014},
  publisher={Dove Press}
}

@article{gsc00,
    author = {Gers, Felix A. and Schmidhuber, Jurgen and Cummins, Fred},
    journal = {Neural Computation},
    number = {10},
    pages = {2451-2471},
    title = {Learning to forget: Continual prediction with LSTM},
    volume = {12},
    year = {2000},
}

@inproceedings{kolesnikov2019revisiting,
  title={Revisiting self-supervised visual representation learning},
  author={Kolesnikov, Alexander and Zhai, Xiaohua and Beyer, Lucas},
  booktitle={Proceedings of the IEEE conference on Computer Vision and Pattern Recognition},
  pages={1920--1929},
  year={2019}
}

@inproceedings{glorot2011domain,
  title={Domain adaptation for large-scale sentiment classification: A deep learning approach},
  author={Glorot, Xavier and Bordes, Antoine and Bengio, Yoshua},
  booktitle={Proceedings of the 28th international conference on machine learning (ICML-11)},
  pages={513--520},
  year={2011}
}

@inproceedings{abadi2016deep,
  title={Deep learning with differential privacy},
  author={Abadi, Martin and Chu, Andy and Goodfellow, Ian and McMahan, H Brendan and Mironov, Ilya and Talwar, Kunal and Zhang, Li},
  booktitle={Proceedings of the 2016 ACM SIGSAC Conference on Computer and Communications Security},
  pages={308--318},
  year={2016}
}

@inproceedings{fredrikson2015model,
  title={Model inversion attacks that exploit confidence information and basic countermeasures},
  author={Fredrikson, Matt and Jha, Somesh and Ristenpart, Thomas},
  booktitle={Proceedings of the 22nd ACM SIGSAC Conference on Computer and Communications Security},
  pages={1322--1333},
  year={2015}
}

@inproceedings{salem2018ecg,
  title={ECG arrhythmia classification using transfer learning from 2-dimensional deep CNN features},
  author={Salem, Milad and Taheri, Shayan and Yuan, Jiann--Shiun},
  booktitle={2018 IEEE Biomedical Circuits and Systems Conference (BioCAS)},
  pages={1--4},
  year={2018},
  organization={IEEE}
}

@article{mathews2018novel,
  title={A novel application of deep learning for single-lead ECG classification},
  author={Mathews, Sherin M and Kambhamettu, Chandra and Barner, Kenneth E},
  journal={Computers in biology and medicine},
  volume={99},
  pages={53--62},
  year={2018},
  publisher={Elsevier}
}

@article{oh2018deep,
  title={A deep learning approach for Parkinson’s disease diagnosis from EEG signals},
  author={Oh, Shu Lih and Hagiwara, Yuki and Raghavendra, U and Yuvaraj, Rajamanickam and Arunkumar, N and Murugappan, M and Acharya, U Rajendra},
  journal={Neural Computing and Applications},
  pages={1--7},
  year={2018},
  publisher={Springer}
}

@article{kim2020predicting,
  title={Predicting the efficiency of prime editing guide RNAs in human cells},
  author={Kim, Hui Kwon and Yu, Goosang and Park, Jinman and Min, Seonwoo and Lee, Sungtae and Yoon, Sungroh and Kim, Hyongbum Henry},
  journal={Nature Biotechnology},
  pages={1--9},
  year={2020},
  publisher={Nature Publishing Group}
}

@inproceedings{penny2020machine,
  title={Machine Learning Algorithms for Predicting and Risk Profiling of Cardiac Surgery-Associated Acute Kidney Injury},
  author={Penny-Dimri, Jahan C and Bergmeir, Christoph and Reid, Christopher M and Williams-Spence, Jenni and Cochrane, Andrew D and Smith, Julian A},
  booktitle={Seminars in Thoracic and Cardiovascular Surgery},
  year={2020},
  organization={Elsevier}
}

@inproceedings{theil2020predicting,
  title={Predicting modality in financial dialogue},
  author={Theil, Kilian and Stuckenschmidt, Heiner},
  booktitle={Proceedings of the 1st Joint Workshop on Financial Narrative Processing and MultiLing Financial Summarisation},
  pages={226--234},
  year={2020}
}

@inproceedings{davis2006relationship,
  title={The relationship between Precision-Recall and ROC curves},
  author={Davis, Jesse and Goadrich, Mark},
  booktitle={Proceedings of the 23rd international conference on Machine learning},
  pages={233--240},
  year={2006}
}

@inproceedings{yosinski2014transferable,
  title={How transferable are features in deep neural networks?},
  author={Yosinski, Jason and Clune, Jeff and Bengio, Yoshua and Lipson, Hod},
  booktitle={Advances in neural information processing systems},
  pages={3320--3328},
  year={2014}
}

@article{chen2016deep,
  title={Deep feature extraction and classification of hyperspectral images based on convolutional neural networks},
  author={Chen, Yushi and Jiang, Hanlu and Li, Chunyang and Jia, Xiuping and Ghamisi, Pedram},
  journal={IEEE Transactions on Geoscience and Remote Sensing},
  volume={54},
  number={10},
  pages={6232--6251},
  year={2016},
  publisher={IEEE}
}

@article{malhotra2017timenet,
  title={TimeNet: Pre-trained deep recurrent neural network for time series classification},
  author={Malhotra, Pankaj and TV, Vishnu and Vig, Lovekesh and Agarwal, Puneet and Shroff, Gautam},
  journal={arXiv preprint arXiv:1706.08838},
  year={2017}
}

@article{rai2016risk,
  title={Risk regulation and innovation: the case of rights-encumbered biomedical data silos},
  author={Rai, Arti K},
  journal={Notre Dame L. Rev.},
  volume={92},
  pages={1641},
  year={2016},
  publisher={HeinOnline}
}

@article{lundberg2020local,
  title={From local explanations to global understanding with explainable AI for trees},
  author={Lundberg, Scott M and Erion, Gabriel and Chen, Hugh and DeGrave, Alex and Prutkin, Jordan M and Nair, Bala and Katz, Ronit and Himmelfarb, Jonathan and Bansal, Nisha and Lee, Su-In},
  journal={Nature machine intelligence},
  volume={2},
  number={1},
  pages={2522--5839},
  year={2020}
}

\clearpage

\section{Supplementary Methods}
\renewcommand{\figurename}{Supplementary Figure}
\renewcommand{\tablename}{Supplementary Table}

\subsection{Data}
\label{sec:app:data}

\subsubsection{Top diagnoses for our data}
\label{sec:app:diagnoses}

Top ten diagnoses (OR$_0$): 

\begin{itemize}
    \itemsep0em 
    \item Cataract NOS
    \item Subarachnoid Hemmorrhage
    \item Calculus of Kidney
    \item Complications due to other internal orthopedic device implant and graft
    \item Senile Cataract NOS
    \item Senile Cataract Unspecified
    \item Necrotizing Fascitis
    \item Cataract
    \item Carpal Tunnel Syndrome
    \item CMP NEC D/T ORTH DEV NEC
\end{itemize}

\noindent
Top ten diagnoses (OR$_1$): 

\begin{itemize}
    \itemsep0em     
    \item Malignant Neoplasm of Breast (Female) Unspecified
    \item Malignant Neoplasm of Breast NOS
    \item Atrial Fibrillation
    \item Morbid Obesity
    \item Calculus of Kidney
    \item Esophageal Reflux
    \item Malignant Neoplasm of Prostate
    \item Malignant Neoplans of Bladder Part Unspecified
    \item PREV C-SECT NOS-DELIVER
    \item End stage renal disease
\end{itemize}

\noindent
Top ten diagnoses (ICU$_\text{P}$): 

\begin{itemize}
    \itemsep0em 
    \item Newborn
    \item Pneumonia
    \item Telemetry
    \item Sepsis
    \item Congestive Heart Failure
    \item Coronary Artery Disease
    \item Chest Pain
    \item Gastrointestinal Bleed
    \item Altered Mental Status
    \item Intracranial Hemorrhage
\end{itemize}

\subsubsection{Labelling}
\label{sec:app:labelling}
For \textit{hypoxemia}, a particular time point $t$ is labelled to be one if the minimum of the next five minutes is hypoxemic ($\min(SAO2^{t+1:t+6})< 93$).  All points where the current time step is currently hypoxemic are ignored ($SAO2^t < 93$).  Additionally we ignore time points where the past ten minutes were all missing or the future five minutes were all missing.  \textit{hypocapnia}, \textit{hypotension}, and \textit{hypertension} have slightly stricter label conditions.  We label the current time point $t$ to be one if ($\min(S^{t-10:t})>T$) and the minimum of the next five minutes is "hypo" ($\min(S^{t+1:t+5})\leq T$).
We label the current time point $t$ to be zero if ($\min(S^{t-10:t})>T$) and the minimum of the next ten minutes is not "hypo" ($\min(S^{t+1:t+10}) > T$).  For \textit{Hypertension}, we use $\max(S^{t+1:t+5})\geq T$ rather than $\min$ and an analogous filtering procedure.  All other time points were not considered.  For \textit{hypocapnia}, the threshold $T=34$ and the signal $S$ is \textit{ETCO2}.  For \textit{hypotension} the threshold $T=59$ and the signal $S$ is 
\textit{NIBPM}.  For \textit{hypertension} the threshold $T=110$ and the signal $S$ is \textit{NIBPM}. Additionally we ignore time points where the past ten minutes were all missing or the future five minutes were all missing.  As a result, we have different sample sizes for different prediction tasks (reported in Table \ref{tab:hospitals}).  For \textit{phenylephrine}, we filter out procedures where phenylephrine is not administered because we would have too many negative samples otherwise.

\subsection{Architecture}
\label{sec:app:arch}

\subsubsection{LSTM (upstream embedding) architecture and training} 
\label{sec:architecture:lstm}

We utilize LSTMs with forget gates, introduced by \cite{gsc00}, implemented in the Keras library with a Tensorflow back-end.  We train our networks with either regression (\textit{auto} and \textit{min} embeddings) or classification (\textit{hypo}) objectives.  For regression, we optimize using Adam with an MSE loss function.  For classification we optimize using RMSProp with a binary cross-entropy loss function (additionally, we upsample to maintain balanced batches during training).  Our model architectures consist of two hidden layers, each with 200 LSTM cells with dense connections between all layers. We found that important steps in training LSTM networks for our data are to impute missing values by the training mean, standardize data, and to randomize sample ordering prior to training. To prevent overfitting, we utilized dropouts between layers as well as recurrent dropouts for the LSTM nodes. We utilized a learning rate of 0.001. Hyperparameter optimization was done by manual coordinate descent.  
The LSTM models were each run for 200 epochs and the final model was selected according to validation loss.   In order to train these models, we utilize three GPUs (GeForce GTX 1080 Ti graphics cards).

\subsubsection{GBM (downstream prediction) architecture and training} 
\label{sec:architecture:gbm}

We train GBM trees in Python using XGB, an open source library for gradient boosting trees.  XGB works well in practice in part due to it's ease of use and flexibility. Imputing and standardizing are unnecessary because GBM trees are based on splits in the training data, so that scale does not matter and missing data is informative as is.  We train the GBM trees with embedding features from 15 physiological signals, resulting in a total of 3000 features for PHASE methods.  In addition, we concatenate static features to the signal features to train and evaluate the models.  We found that a learning rate of 0.02 for hypoxemia (0.1 for hypotension and hypocapnia), a max tree depth of 6, subsampling rate of 0.5, and a logistic objective gave us good performance.  We fix hyperparameter settings across experiments so that we can focus on comparing different representations our signal data.  All XGB models were run until their validation accuracy was non-improving for five rounds of adding estimators (trees).  In order to train these models, we utilize 72 CPUs (Intel(R) Xeon(R) CPU E5-2699 v3 @ 2.30GHz)

\section{Supplementary Results}

\subsection{Results in AP and ROCAUC scale}
\label{sec:app:absolute_AP}

In this section, we report the AP scores for each of our main text plots Figures (\ref{tab:ap_first}-\ref{tab:ap_last}).  In the main text we report the transfer AP.  Additionally we report a shorter version of the performance/transference results for hypo outcomes in ROC AUC in Figure \ref{fig:roc_short}.

\begin{table}[htbp]
\centering
\begin{tabular}{llccc}
Downstream Model                 &  Data Type    & Hypoxemia & Hypocapnia & Hypotension \\ \hline
LSTM                 & \textit{raw}  & 0.2431    & 0.4109     & 0.2898      \\ \hline
\multirow{7}{*}{XGB} & \textit{raw}  & 0.2411    & 0.4243     & 0.2996      \\
                     & \textit{ema}  & 0.2420    & 0.4277     & 0.2890      \\
                     & \textit{rand} & 0.2135    & 0.3973     & 0.2630      \\
                     & \textit{auto} & 0.2402    & 0.4246     & 0.2974      \\
                     & \textit{next} & 0.2585    & 0.4426     & 0.3116      \\
                     & \textit{min}  & 0.2621    & 0.4392     & 0.3127      \\
                     & \textit{hypo} & 0.2581    & 0.4471     & 0.3100     
\end{tabular}
\caption{Non-transference results for hypo outcomes in AP, corresponding to Figure \ref{fig:performance}b.}
\label{tab:ap_first}
\end{table}

\begin{table}[htbp]
\centering
\begin{tabular}{lccc}
Data Type     & Hypoxemia & Hypocapnia & Hypotension \\
\textit{next}     & 0.2585    & 0.4426     & 0.3116      \\
\textit{next}$'$  & 0.2529    & 0.4451     & 0.3123      \\
\textit{next}$^P$ & 0.2486    & 0.4243     & 0.2996      \\
\textit{min}      & 0.2621    & 0.4392     & 0.3127      \\
\textit{min}$'$   & 0.2580    & 0.4414     & 0.3140      \\
\textit{min}$^P$  & 0.2568    & 0.4243     & 0.2996      \\
\textit{hypo}     & 0.2581    & 0.4471     & 0.3100      \\
\textit{hypo}$'$  & 0.2499    & 0.4342     & 0.2879      \\
\textit{hypo}$^P$ & 0.2383    & 0.4243     & 0.2996     
\end{tabular}
\caption{Transference results for hypo outcomes in AP, corresponding to Figure \ref{fig:performance}c.}
\end{table}

\begin{table}[htbp]
\centering
\begin{tabular}{lccc}
Data Type                       & Hypoxemia & Hypocapnia & Hypotension \\
\textit{next}         & 0.2585    & 0.4426     & 0.3116      \\
\textit{next}$'$      & 0.2529    & 0.4451     & 0.3123      \\
\textit{next}         & 0.2523    & 0.4315     & 0.2829      \\
\textit{next}$'_{m}$ & 0.2486    & 0.4234     & 0.2879     
\end{tabular}
\caption{Jointly trained model results in AP, corresponding to Figure \ref{fig:joint_performance}.}
\end{table}

\begin{table}[htbp]
\centering
\begin{tabular}{cccc|ccc|ccc}
\multirow{2}{*}{} & \multicolumn{3}{c}{Hypoxemia}                      & \multicolumn{3}{c}{Hypocapnia}                     & \multicolumn{3}{c}{Hypotension}                    \\
\# Signals              & \textit{raw} & \textit{next} & \textit{next}$'$ & \textit{raw} & \textit{next} & \textit{next}$'$ & \textit{raw} & \textit{next} & \textit{next}$'$ \\
1  & 0.2413        & 0.2467         & 0.2393            & 0.3950        & 0.4006         & 0.4009            & 0.2590        & 0.2576         & 0.2552            \\
3  & 0.2412        & 0.2535         & 0.2469            & 0.4109        & 0.4151         & 0.4149            & 0.2722        & 0.2771         & 0.2741            \\
5  & 0.2402        & 0.2559         & 0.2505            & 0.4256        & 0.4307         & 0.4312            & 0.2828        & 0.3005         & 0.2960            \\
7  & 0.2417        & 0.2561         & 0.2522            & 0.4275        & 0.4251         & 0.4319            & 0.2948        & 0.3123         & 0.3091            \\
9  & 0.2414        & 0.2572         & 0.2511            & 0.4273        & 0.4303         & 0.4352            & 0.2952        & 0.3088         & 0.3077            \\
11 & 0.2393        & 0.2568         & 0.2518            & 0.4272        & 0.4314         & 0.4390            & 0.2969        & 0.3130         & 0.3105            \\
13 & 0.2414        & 0.2556         & 0.2492            & 0.4251        & 0.4337         & 0.4386            & 0.2943        & 0.3125         & 0.3079            \\
15 & 0.2419        & 0.2565         & 0.2508            & 0.4224        & 0.4346         & 0.4410            & 0.2943 & 0.3095   &	0.3073
\end{tabular}
\caption{Heterogeneous results in AP, corresponding to Figure \ref{fig:heterogeneous}.}
\end{table}

\begin{table}[htbp]
\centering
\begin{tabular}{lcc}
Data Type        & Hypertension & Phenylephrine \\
raw     & 0.2275       & 0.1613        \\
ema     & 0.2227       & 0.1569        \\
auto    & 0.2270       & 0.1617        \\
auto$'$ & 0.2271       & 0.1584        \\
next    & 0.2340       & 0.1703        \\
next$'$ & 0.2349       & 0.1674        \\
min     & 0.2355       & 0.1665        \\
min$'$  & 0.2329       & 0.1662       
\end{tabular}
\caption{Non-hypo outcome results in AP, corresponding to Figure \ref{fig:performance}d.}
\end{table}

\begin{table}[htbp]
\centering
\begin{tabular}{lccc}
Outcome       & next   & next$'$ & next$^\text{ft}$ \\
Hypoxemia     & 0.2585 & 0.2529  & 0.2597          \\
Hypocapnia    & 0.4426 & 0.4451  & 0.4447          \\
Hypotension   & 0.3116 & 0.3123  & 0.3143          \\
Hypertension  & 0.1703 & 0.1674  & 0.1715          \\
Phenylephrine & 0.2340 & 0.2349  & 0.2386         
\end{tabular}
\caption{Fine tuning results in AP, corresponding to Figure \ref{fig:ft_performance}.}
\label{tab:ap_last}
\end{table}

\begin{table}[htbp]
\centering
\begin{tabular}{lcc|cc|cc}
 & \multicolumn{2}{c}{Hypoxemia}                      & \multicolumn{2}{c}{Hypocapnia}                     & \multicolumn{2}{c}{Hypotension}                    \\
Data Type            & Mean    & SE             & Mean    & SE             & Mean    & SE             \\
\textit{raw}         & 0.88665 & 0.00019        & 0.84249 & 0.00016        & 0.87122 & 0.00015        \\
\textit{ema}         & 0.89161 & 0.00017        & 0.84663 & 0.00016        & 0.87002 & 0.00014        \\
\textit{auto}        & 0.89085 & 0.00018        & 0.84652 & 0.00016        & 0.87445 & 0.00013        \\
\textit{next}    & 0.89340 & 0.00017        & 0.85284 & 0.00016        & 0.87917 & 0.00014        \\
\textit{min}        & 0.89530 & 0.00017        & 0.85240 & 0.00016        & 0.88056 & 0.00014        \\
\textit{hypo}        & 0.89398 & 0.00017        & 0.85509 & 0.00016        & 0.87826 & 0.00014        \\
\textit{next}$'$ & 0.89216 & 0.00017        & 0.85328 & 0.00016        & 0.87921 & 0.00014        \\
\textit{min}$'$     & 0.89378 & 0.00017        & 0.85387 & 0.00016        & 0.88072 & 0.00013        \\
\textit{hypo}$'$     & 0.89115 & 0.00017        & 0.84914 & 0.00016        & 0.86964 & 0.00014       
\end{tabular}
\caption{Shorter version of performance/transference results for hypo outcomes in ROC AUC, corresponding to Figures \ref{fig:performance}b-c.}
\label{fig:roc_short}
\end{table}

\subsubsection{Benchmarking against a jointly trained embedding model}
\label{sec:app:joint}

\begin{figure}[htbp]
    \centering
    \includegraphics[width=.8\textwidth]{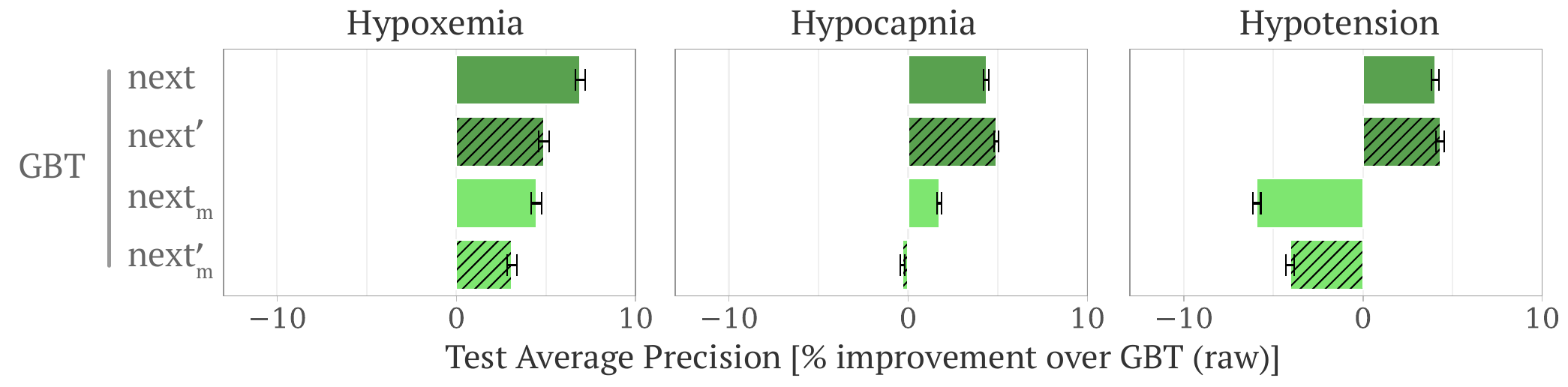}
    \caption{\textit{Performance of a jointly trained model}.  We train a downstream XGB model to forecast adverse outcomes with different representations of data.  First, \textit{next} is the same as in Figure \ref{fig:performance}.  Then, \textit{next$_\text{m}$} is our jointly trained model.  Rather than training each of the 15 LSTMs separately, we train a larger model that jointly trains all 15 networks to forecast the next five minutes of all 15 signals.  Finally, ($'$) denotes transference, as in Figure \ref{fig:performance}.}
    \label{fig:joint_performance}
\end{figure}

In this section, our aim is to investigate whether jointly training the embedding models affords any benefits over training per-signal networks (more details in Section \ref{sec:meth:joint}).  To do so, we compare the per-signal LSTM embedding models to a much larger LSTM model that is jointly trained.  In order to ensure a fair comparison, we used exactly the same hyperparameters (batch size, learning rate, optimizer, number of epochs, etc.) and used roughly the same number of parameters in both cases.  For the per-signal models we used 2 layers of 200 LSTM nodes to predict the next five minutes of a given signal (\textit{next}).  In the larger LSTM model, we take 15 of these 2 layer networks and concatenate their outputs which is used to predict 15 distinct tasks: the next five minutes of each signal (minimizing mean squared error for all tasks simultaneously).  Then we use this jointly trained model to create embeddings for each signal (\textit{next}$_\text{m}$).

In terms of the performance of XGB trained on these embeddings, we observe that \textit{next$_\text{m}$} performs almost as well as the per-signal \textit{next} models.  However, for Hypocapnia and Hypotension, it does not perform nearly as well, even doing worse than \textit{raw} for Hypotension.  This could be due to a number of reasons.  One possible reason is that multi-task learning is usually beneficial because it increases the effective sample size for one specific outcome using related outcomes; however, in this case we have a large enough sample size to learn the best possible representations.  Another possible reason is that the size of the network is much larger in the jointly learned LSTM.  With the addition of a loss function that includes many separate signals it may simply constitute a harder optimization problem.  As such, we recommend utilizing per-signal embedding models which confer a number of other benefits including using less GPU memory and converging much faster, unless there is a strong reason to do otherwise.

\subsection{MLP downstream model}
\label{sec:app:mlp_performance}

\begin{figure}[htbp]
    \centering
    \includegraphics[width=\textwidth]{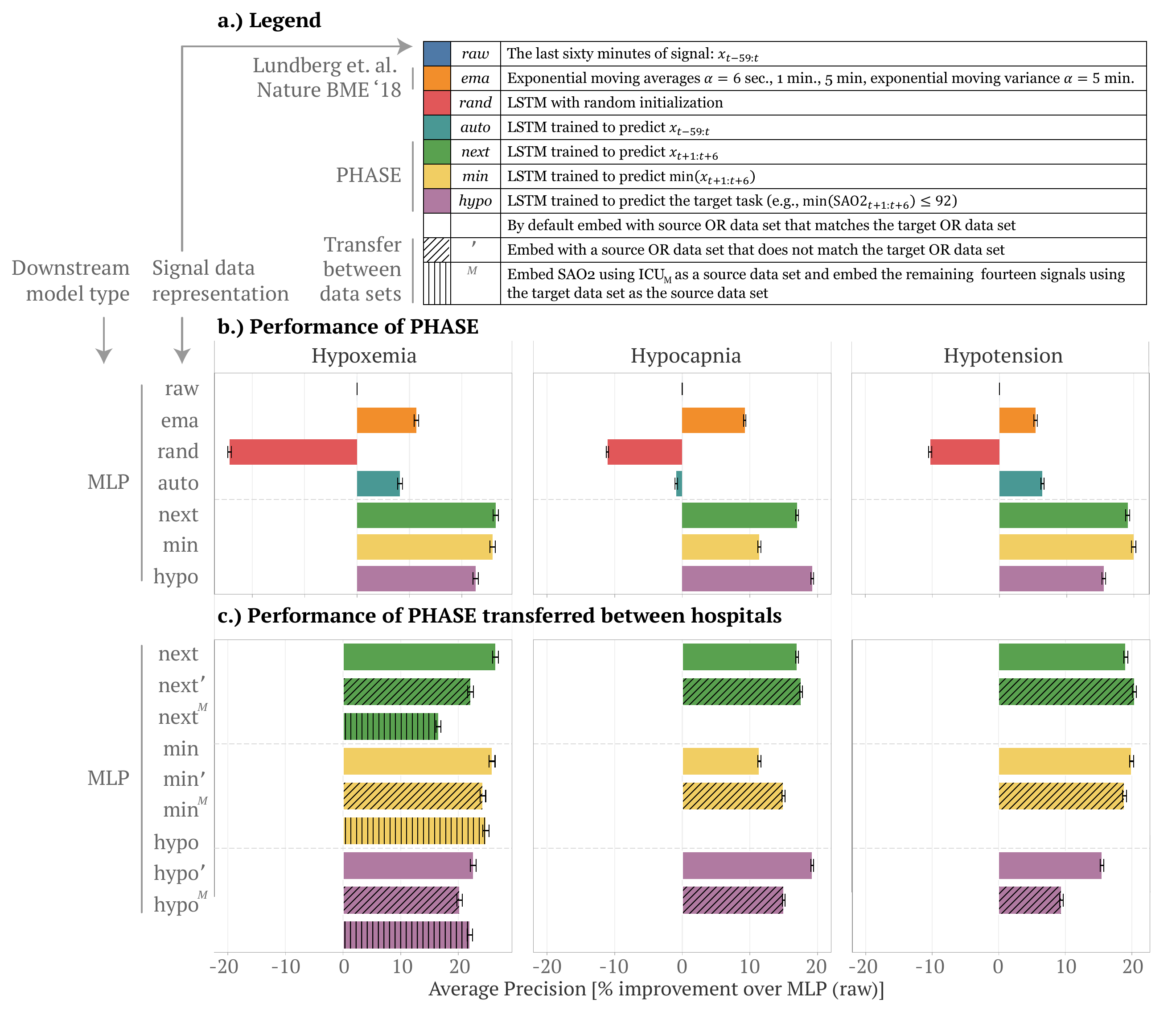}
    \caption{Performance of PHASE embeddings with MLP downstream model rather than XGB as in Figure \ref{fig:performance}.}
    \label{fig:mlp_performance}
\end{figure}

One potential criticism of our evaluations in previous are that we evaluate with only one downstream model type: XGB.  There are clearly a number of benefits to using tree-based methods such as ease of training, exact SHAP value attribution methods, performance on par with LSTMs for our data sets, and more.  However, in order to show that \textit{PHASE embeddings improve performance and transference for a variety of downstream model types}, we replicate Figure \ref{fig:performance}b and \ref{fig:performance}c using MLP downstream models in Figure \ref{fig:mlp_performance}.  We see that, as with downstream XGB models, the PHASE embeddings offer a substantial improvement over \textit{raw} embeddings for downstream MLP models.

We utilize multi-layer perceptrons, implemented in the Keras library with a Tensorflow back-end.  We train the MLPs with embedding features from 15 physiological signals, resulting in a total of 3000 features for PHASE methods.  In addition, we concatenate static features to the signal features to train and evaluate the models.  The model's architecture consists of the following: a dense layer with 100 nodes (with a relu activation) followed by a dropout layer with dropout rate 0.5 followed by a dense layer with 100 nodes (with a relu activation) followed by a dropout layer with dropout rate 0.5 followed by the dense output layer with one node and sigmoid activation function.  We utilize a learning rate of 0.00001, adam optimizer, and binary cross entropy loss.  We found that 200 epochs was sufficient for the downstream models to converge.  We fix hyperparameter settings across experiments so that we can focus on comparing different representations our signal data.  In order to train these models, we utilize 72 CPUs (Intel(R) Xeon(R) CPU E5-2699 v3 @ 2.30GHz)

\subsection{Applying PHASE for heterogeneous features}
\label{sec:app:heterogeneous}

\begin{figure}[htbp]
    \centering
    \includegraphics[width=.9\textwidth]{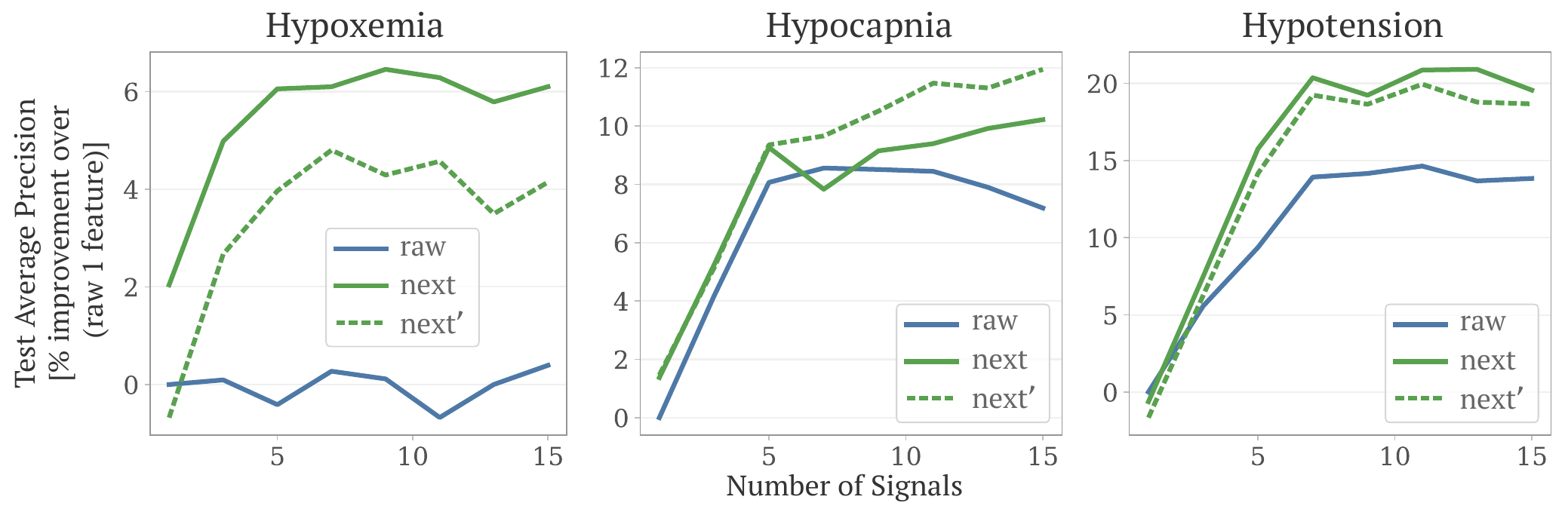}
    \caption{\textit{Heterogeneous features.}  Performance of PHASE for XGB trained on subsets of the features on the target data.  For these models we do not concatenate the six static features and instead focus on the signal data exclusively.}
    \label{fig:heterogeneous}
\end{figure}

In this section, we aim to evaluate PHASE for heterogeneous feature sets by acting as if the target data has fewer physiological signals than it actually does.  To do so, we impose an ordering on the features according to the importance of each signal for the XGB model trained on \textit{raw} data (Supplementary Figures \ref{fig:hypoxemia_attr}b for hypoxemia, \ref{fig:hypocapnia_attr}b for hypocapnia, \ref{fig:hypotension_attr}b for hypotension).  Then, we run XGB to forecast our outcomes with increasing subsets of features by including the most important features first.  In Figure \ref{fig:heterogeneous}, we can see that for all three outcomes, \textit{next} and \textit{next}$'$, are either comparable to the raw representation or significantly better.  One interesting relationship is that the improvement over the \textit{raw} signal becomes greater with an increased number of signals.  This implies that the embeddings allow the downstream model to better use signals in conjunction, whereas the embedding of the most important feature for each task is not as useful on its own.

\subsection{Full summary plots}
\label{sec:app:full_attr}

\begin{figure}[htbp]
    \centering
    \includegraphics[width=\textwidth]{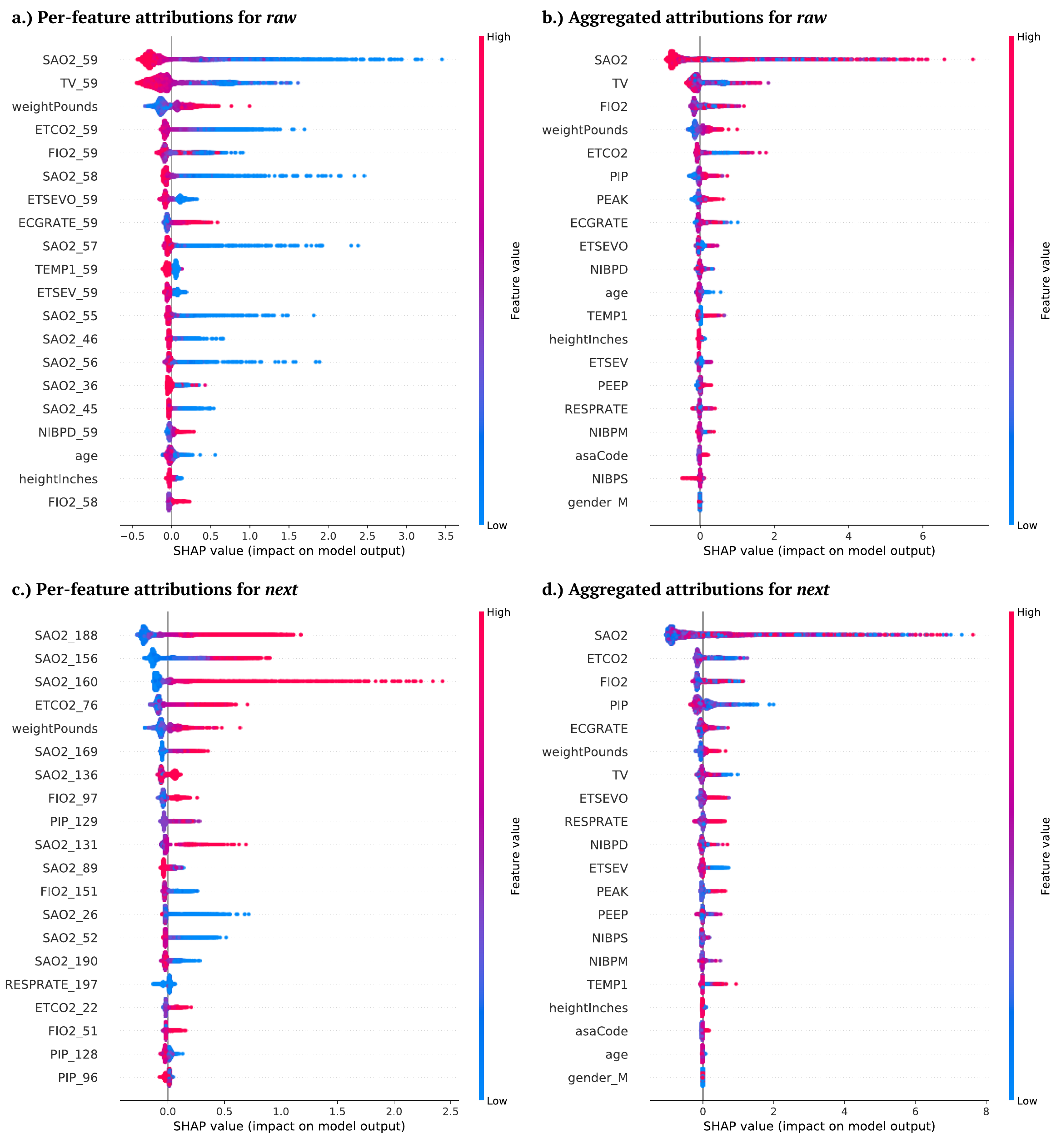}
    \caption{Attributions for hypoxemia.}
    \label{fig:hypoxemia_attr}
\end{figure}

\begin{figure}[htbp]
    \centering
    \includegraphics[width=\textwidth]{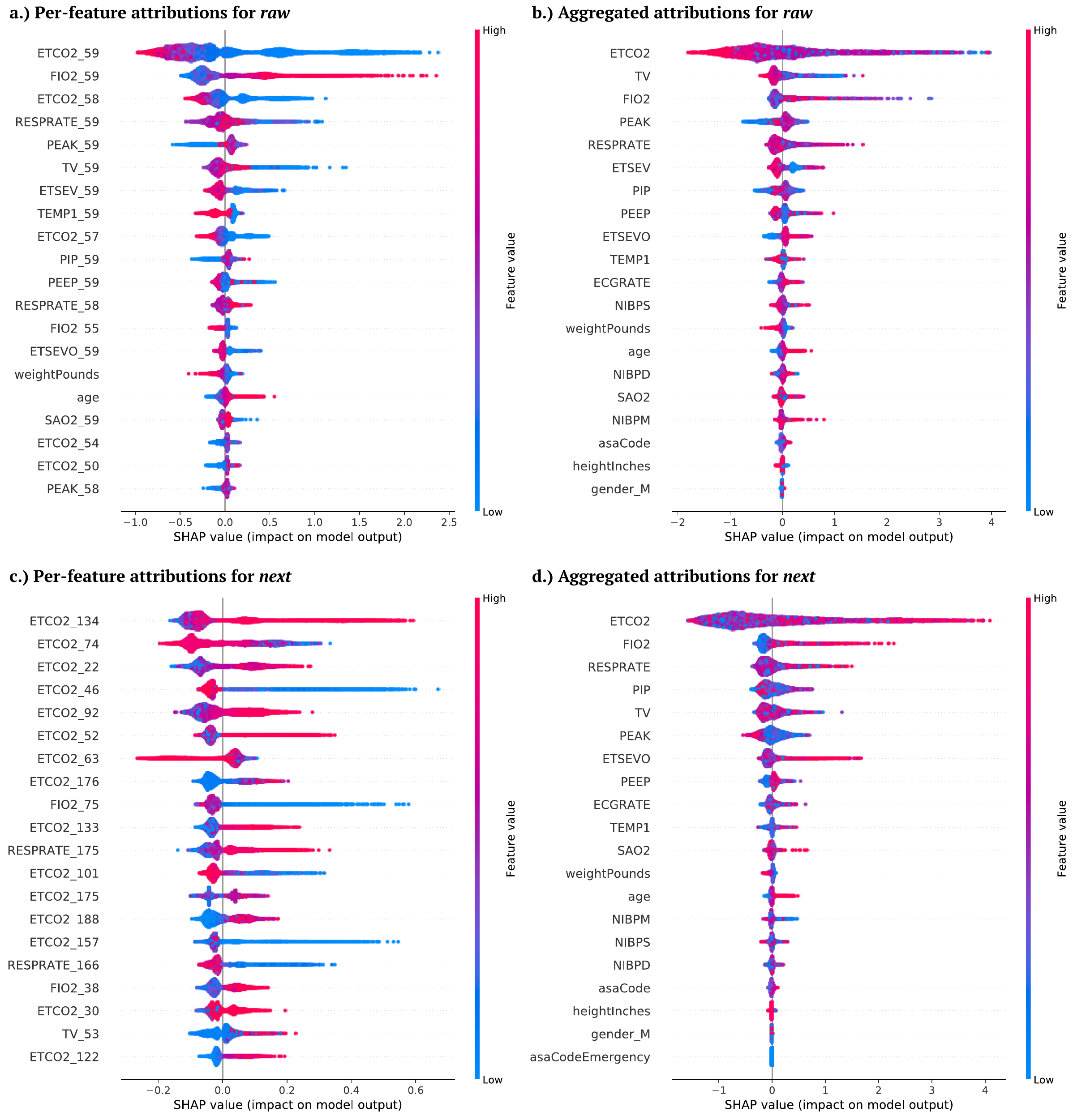}
    \caption{Attributions for \textit{hypocapnia}.}
    \label{fig:hypocapnia_attr}
\end{figure}

\begin{figure}[htbp]
    \centering
    \includegraphics[width=\textwidth]{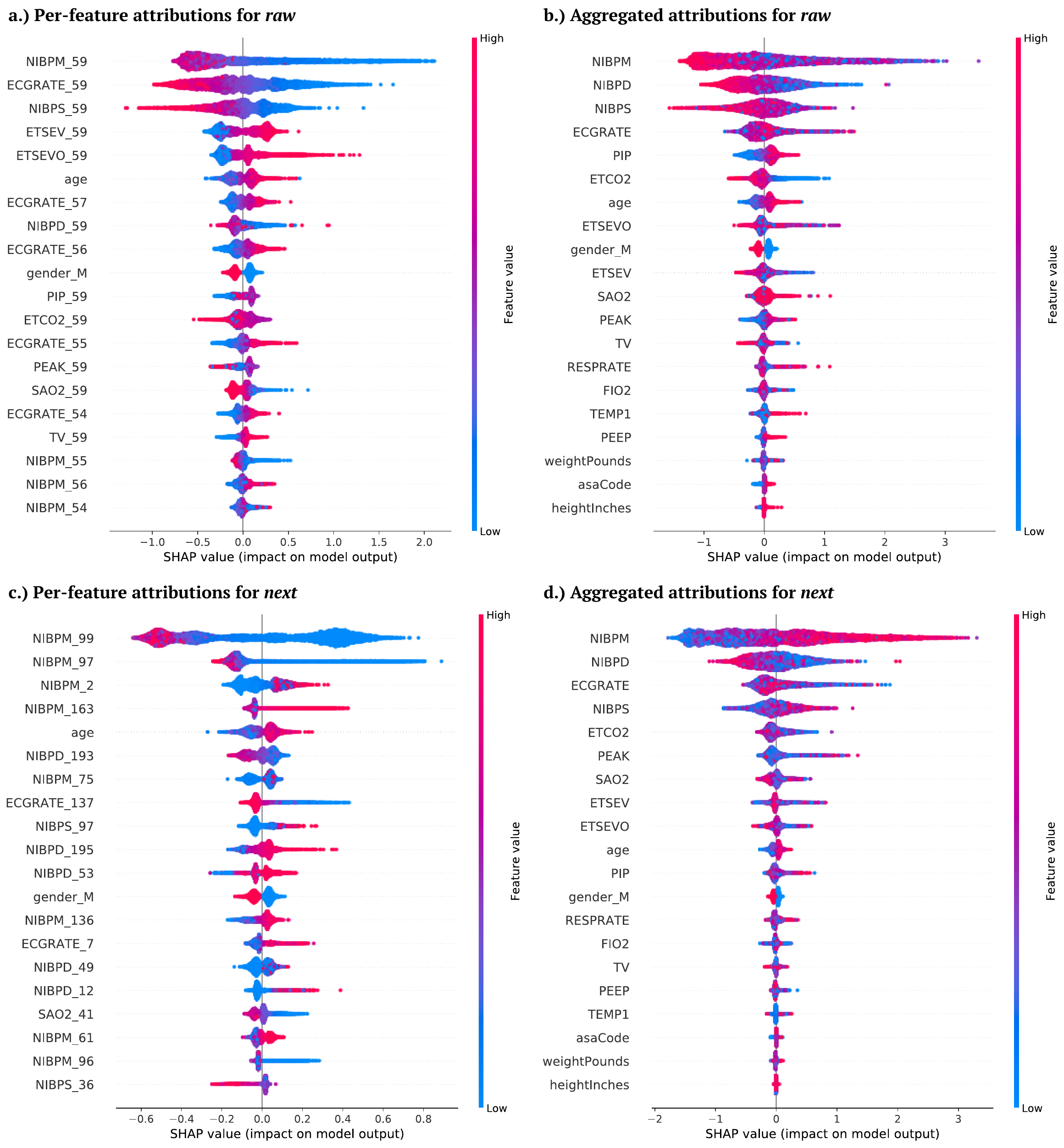}
    \caption{Attributions for \textit{hypotension}.}
    \label{fig:hypotension_attr}
\end{figure}

\begin{figure}[htbp]
    \centering
    \includegraphics[width=\textwidth]{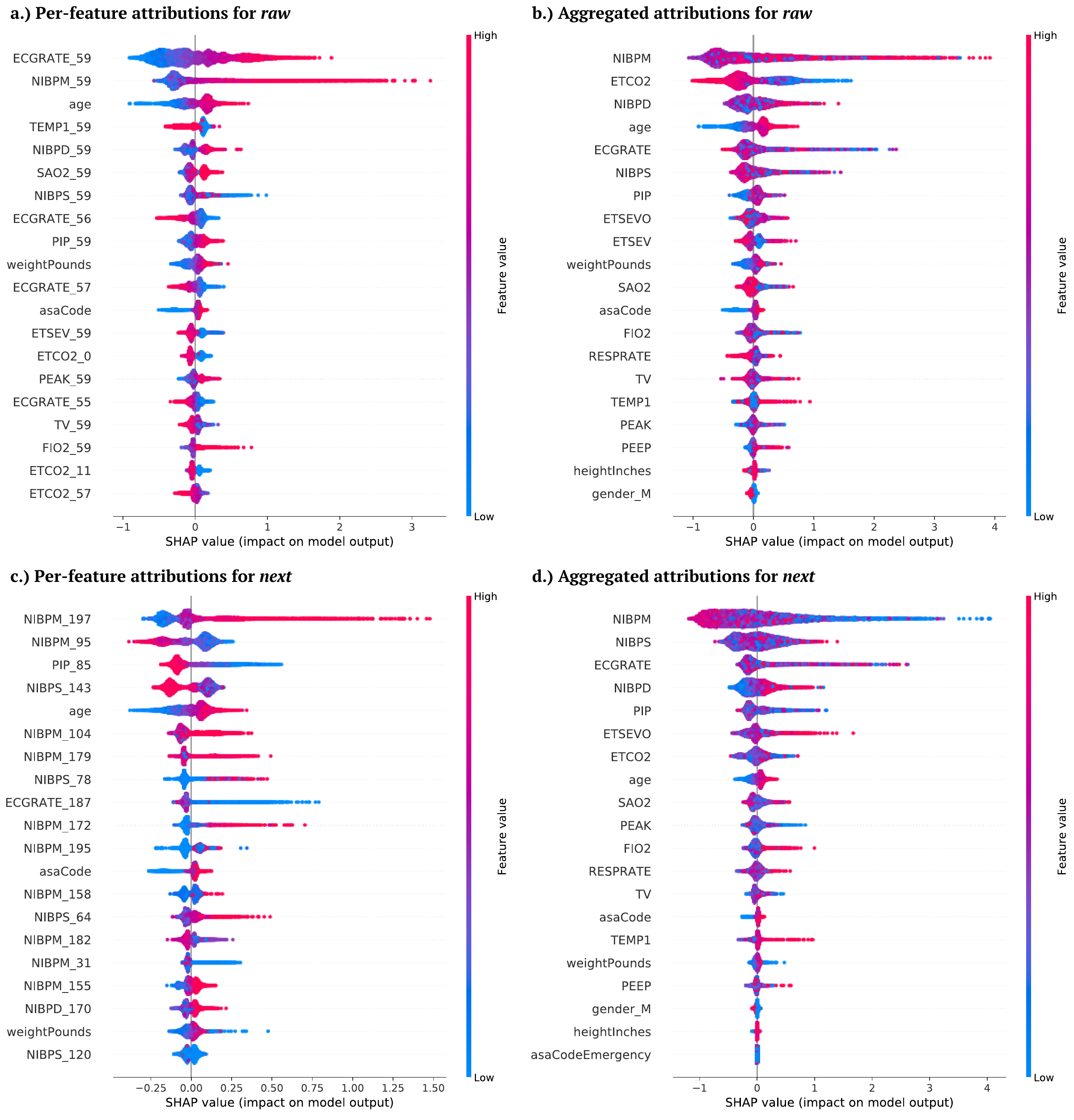}
    \caption{Attributions for \textit{hypertension}.}
    \label{fig:hypertension_attr}
\end{figure}

\begin{figure}[htbp]
    \centering
    \includegraphics[width=\textwidth]{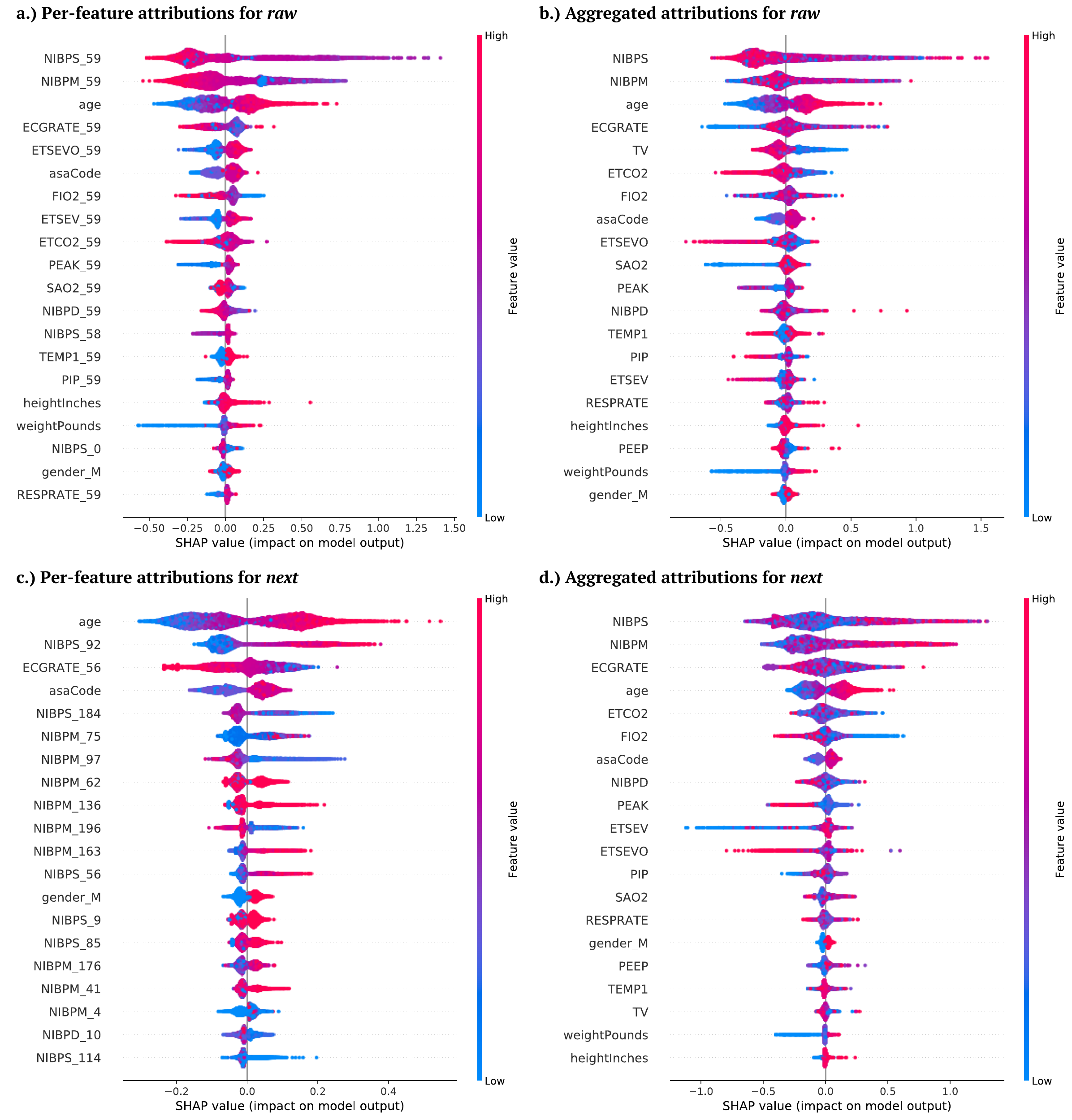}
    \caption{Attributions for \textit{phenylephrine}.}
    \label{fig:phenylephrine_attr}
\end{figure}

In this section, we show the full summary plots (Figures \ref{fig:hypoxemia_attr}-\ref{fig:phenylephrine_attr}) for the per-feature and aggregated attributions for \textit{raw} and \textit{next} models trained from XGB models trained in target data set OR$_0$.  We can see more relationships between each of the five downstream tasks and the top 20 features sorted by the mean absolute SHAP values for each feature.

\end{document}